\documentclass{article}

\usepackage{PRIMEarxiv}

\usepackage[utf8]{inputenc} 
\usepackage[T1]{fontenc}    
\usepackage{hyperref}       
\usepackage{url}            
\usepackage{booktabs}       
\usepackage{amsfonts}       
\usepackage{nicefrac}       
\usepackage{microtype}      
\usepackage{lipsum}
\usepackage{fancyhdr}       
\usepackage{graphicx}       
\usepackage{amsmath}        
\usepackage{amssymb}        
\usepackage{enumitem}       
\usepackage{caption}
\usepackage{subcaption}
\usepackage{multirow}
\usepackage{tikz}

\usetikzlibrary{arrows.meta}
\graphicspath{{media/}}     

\usepackage{acro}   
\DeclareAcronym{sr}{
  short       = SR ,
  long        = symbolic regression ,
  short-plural =  ,
  long-plural  = 
}
\DeclareAcronym{llm}{
  short       = LLM ,
  long        = large language model ,
  short-plural = s,
  long-plural  = s
}

\newcommand{\automodel}{\texttt{automodel}}
  
\title{\texttt{automodel}: A Skill for Agentic Model Discovery
}

\author{
  Marco Virgolin \\
  Unlayer AI \\
  Amsterdam, the Netherlands \\
  \texttt{marco@unlayer.ai} \\
    \And
  Harry Desmond \\
  Institute of Cosmology and Gravitation \\
  University of Portsmouth \\
  Portsmouth, UK \\
  \texttt{harry.desmond@port.ac.uk} \\
   \And
   Ester Bergantin \\
   Department of Mathematics \\
   University of Trento \\ 
   Trento, Italy \\
   \texttt{ester.bergantin@unitn.it} \\
   \And
  Giorgio Morales\\
  Aston Centre for Artificial Intelligence\\ Research and Application,\\ Aston University, \\
  Birmingham, UK \\
  \texttt{giorgiomorales@ieee.org} \\
   \And
  John W. Sheppard \\
  Gianforte School of Computing,\\ Montana State University, \\
  Bozeman, USA \\
  \texttt{john.sheppard@montana.edu} \\
   \And
  Luigi Romano \\
  Department of Electrical Engineering \\
  Linköping University \\
  Linköping, Sweden \\
  \texttt{luigi.romano@liu.se} \\
   \And
  Alessandro Lucantonio \\
Department of Mechanical \\ and Production Engineering, \\
  Aarhus University, \\
  Aarhus, Denmark \\
  \texttt{a.lucantonio@mpe.au.dk} \\
}

\begin{document}
\maketitle

\begin{abstract}
The search for better models, whether expressed as analytical equations or neural architectures, has increasingly shifted from manual design toward data-driven automation.
While a multitude of algorithms exist, ranging from exhaustive search to AlphaEvolve, recent advances in general-purpose coding agents such as Claude Code and OpenAI Codex raise the question: could providing such systems with a ``model discovery recipe'' be enough to discover or improve models?
We present \automodel, an agent skill for general model discovery.
With case studies in astrophysics, cardiovascular dynamics, precision agriculture, traffic flow, and vehicle tire dynamics, we show how commercial agents equipped with \automodel{} can successfully identify novel and scientifically-relevant models.
Our findings suggest that encoding model-discovery methodology as a portable agent skill can provide a simple, interactive, powerful and broadly applicable alternative to domain-specific automated discovery pipelines.
The skill is available at: \url{https://github.com/unlayer-ai/automodel}.

\end{abstract}


\section{Introduction}
For centuries, the modeling of phenomena has relied on human expertise.
Since the 20\textsuperscript{th} century, advances in computational power and algorithms have made it possible to automate many steps in this process.
Heuristics for discovering models as physics laws from data emerged before the 1990s~\cite{langley1987scientific},  predating the subsequent development of \emph{genetic programming}-based \ac{sr}, where compositions of operators, variables, and constants that fit the data are evolved~\cite{koza1992genetic,kronberger2024symbolic}.
In 2009, Schmidt and Lipson used \ac{sr} to demonstrate that Hamiltonians, Lagrangians, and other known physical laws could be recovered automatically from observational data~\cite{schmidt2009distilling}.
Today, modern \ac{sr} algorithms range from evolutionary to deep learning approaches, and have proven capable of discovering \emph{new} candidate laws, for example in astrophysics~\cite{bartlett2024,cranmer2020discovering}. 

Models need not take the form of human-interpretable equations.
\emph{Neural architecture search} (NAS) seeks to automate the design of neural network architectures~\cite{zoph2017neural}, while the broader field of \emph{automated machine learning} (AutoML) also addresses the selection of model classes, preprocessing steps, and hyperparameter configurations~\cite{feurer2015efficient,he2021automl}.
Although these tasks target different model representations, they often draw on the same families of search algorithms, including evolutionary algorithms, Bayesian optimization, and (deep) reinforcement learning.
At a high level, they involve searching over structural choices---such as mathematical operators, neural network layers, or model classes---while also optimizing parameters such as equation coefficients, network weights, and regularization strengths.
Crucially, the structural choices and allowed combinations need to be pre-specified:
The modeler relying on these algorithms must therefore put effort into adapting and configuring these algorithms to the problem and data at hand.

Recent advances in \acp{llm} have expanded the possibilities for automated model discovery.
Systems such as \emph{AlphaEvolve}~\cite{novikov2025alphaevolve}, \emph{ShinkaEvolve}~\cite{lange2026shinkaevolve}, \emph{OpenEvolve}~\cite{openevolve}, and \emph{SkyDiscover}~\cite{liu2026skydiscover} use \acp{llm} to generate or revise executable programs based on task descriptions and evaluation feedback.
When these programs implement models, this becomes a search over model structures.
LLM-generated proposals are generally more expensive than simple mutations, but can incorporate domain knowledge and evidence from previous attempts; their added cost may therefore be offset by fewer evaluations needed to find useful improvements.
This makes LLM-guided search particularly attractive when evaluating candidates requires expensive training, calibration, or simulation.
Further, representing models directly as programs also reduces the need for model-specific encodings and variation operators.

The aforementioned methods rely on dedicated \emph{harnesses}---software that coordinates LLM calls, candidate generation, and evaluation---to organise and scale discovery.
Meanwhile, general-purpose coding agents such as Codex~\cite{openaiCodex}, Claude Code~\cite{anthropicClaudeCode}, and OpenCode~\cite{opencode} can inspect code, modify implementations, execute experiments, and use the results to guide subsequent changes.
These capabilities motivate a complementary approach: can model discovery be organised as a reusable scientific workflow executed by a general-purpose coding agent?

In this work, we investigate this idea through \automodel{}, a portable \emph{skill} we introduce that provides coding agents with a reusable recipe for model discovery.
The skill guides the agent through establishing an evaluator, proposing and implementing hypotheses, conducting experiments, and verifying apparent improvements.
Models are assessed against objectives and scientific constraints specified by the researcher, while their representation and proposed modifications are adapted to the problem and the existing codebase.
This allows the same workflow to address tasks ranging from discovering compact analytical expressions to modifying equations within scientific simulators.

The contribution posed in this paper is methodological: we package model discovery as a reusable skill that operates within existing coding-agent environments, supporting interactive problem formulation as well as model improvement.
Through five scientific case studies, we examine its applicability across model representations, objectives, and validation requirements.
Our aim is to assess the utility and limitations of this agentic workflow---in which researchers can interactively refine objectives, constraints, and evaluation procedures---rather than attempt to establish performance superiority over dedicated model-discovery methods.

\section{Related Work}
As mentioned in the introduction, model discovery spans a number of algorithms and model forms, from genetic programming for \ac{sr}~\cite{kronberger2024symbolic} to AutoML pipelines~\cite{feurer2015efficient,he2021automl} and \emph{LLM-based approaches}.
Here, we focus on the latter since \automodel{} is a skill for LLM agents.

\paragraph{LLMs as optimizers and program generators.}
In \emph{Large Language Models as Optimizers}, Yang et al.\ introduced Optimization by PROmpting (OPRO), in which an LLM is iteratively prompted to create a solution attempt, based on its previous attempts and respective scores, which are evaluated by an objective function~\cite{yang2024opro}.
While not applied to model discovery, OPRO's pattern of proposal, evaluation, and feedback represents the core principle of LLM-based discovery.
FunSearch applies this principle to executable programs, combining LLM-generated candidates with automated evaluation and a population-based search to discover mathematical constructions and algorithmic heuristics~\cite{romeraparedes2024funsearch}.
AlphaEvolve, ShinkaEvolve, OpenEvolve, and SkyDiscover broaden evolutionary code improvement to scientific and computational applications~\cite{novikov2025alphaevolve, lange2026shinkaevolve,openevolve,liu2026skydiscover}.
AlphaEvolve in particular has been used to search for neural-network activation functions, while EvoPrompting has applied LLM-guided evolution to code-level NAS~\cite{vitvitskyi2026activations,chen2023evoprompting}.


\paragraph{From equation proposals to scientific agents.}
LLM-SR specializes in equation discovery, alternating LLM-generated equation structures with numerical parameter fitting and evolutionary selection~\cite{shojaee2025llmsr}.
OpenEvolve has been evaluated on \ac{sr} for three-dimensional surface equations~\cite{castro2025cogfunsearch,kabra2026surfacebench}.
For dynamical systems, LLM4ED uses scored proposals and LLM-based evolutionary operators to discover governing ordinary and partial differential equations~\cite{du2024llm4ed}.
LEMMA evolves mechanistic ecosystem models while fitting their parameters to time-series data~\cite{spillias2026lemma}.
LaSR guides symbolic search using a library of textual concepts extracted from successful hypotheses~\cite{grayeli2024lasr}.
SR-Scientist proposes a harness that allows us to analyze data, implement equations, invoke evaluation tools, and refine candidates from experimental feedback~\cite{xia2026srscientist}.

\paragraph{Autoresearch and reusable experiment loops.}
Karpathy's \texttt{autoresearch} provides a particularly close software precedent to \automodel~\cite{karpathy2026autoresearch}.
While the works above propose harnesses that are specialized to scalable search, Karpathy takes a general coding agent (e.g., OpenAI Codex or Claude Code) and provides a set of instructions in a Markdown file, plus a series of scripts to realize the experiment (improve language-model training code).
This idea has since been generalized into a skill featuring user-defined goals, editable scope, verification procedures, baseline measurement, and persistent result logging~\cite{goenka2026autoresearch}.
Essentially, \texttt{autoresearch} showed how natural-language procedures and a small amount of supporting code infrastructure can lead to sustained optimization within existing generalist coding-agent environments.

\paragraph{Skills for research workflows.}
Skills have also been proposed for broader research methodology.
ARIS coordinates research execution and cross-model review through Markdown-defined skills, persistent research records, and checks linking scientific claims to experimental evidence~\cite{yang2026aris}. DisCo, introduced in \emph{Repo-To-Skill}, distills operational knowledge from machine-learning repositories into reusable skills and supplies them to a research agent~\cite{chen2026repotoskill}.

\automodel{} belongs to this emerging family of instruction-driven research workflows.
It packages a methodology specifically for scientific model discovery: establishing an evaluator, fitting a baseline, iterating over structural hypotheses with repeated parameter calibration, and performing an application-specific final assessment.

\section{Method}

\automodel{} is an agent skill: a set of natural-language instructions and templates arranged in a specific format, for general-purpose coding agents to follow. 
It specifies a workflow rather than providing a dedicated orchestration engine or numerical optimizers; execution relies on the coding agent and its environment.
The skill guides the agent to collaborate with the user to define the search problem, followed by autonomous optimization of model structure.
The type of model is not pre-specified: it may as well be a closed-form expression as a deep neural network.

\begin{figure}[!h]
\centering
\begin{tikzpicture}[
    >={Latex[length=2mm]},
    phase/.style={draw=black!65, rounded corners=2pt, align=center,
      text width=2.55cm, minimum height=1.32cm, inner sep=4pt,
      font=\small},
    inner/.style={draw=black!55, rounded corners=2pt, align=center,
      text width=2.95cm, minimum height=0.86cm, inner sep=3pt,
      font=\small},
    flow/.style={->, thick, draw=black!70}
]
\node[phase, fill=blue!5] (setup) at (0,0)
  {\textbf{1. Setup}\\Goal, data, score, fitting routine};
\node[phase, fill=blue!5] (baseline) at (3.35,0)
  {\textbf{2. Baseline}\\Fit model; verify evaluator};
\node[phase, fill=green!8] (iterate) at (6.70,0)
  {\textbf{3. Iterate}\\Change structure; fit candidates};
\node[phase, fill=blue!5] (finalize) at (10.05,0)
  {\textbf{4. Finalize}\\Final assessment};
\draw[flow] (setup.east) -- (baseline.west);
\draw[flow] (baseline.east) -- (iterate.west);
\draw[flow] (iterate.east) -- (finalize.west);

\draw[dashed, rounded corners=3pt, draw=black!50]
  (-1.43,-1.83) rectangle (11.49,-4.73);
\node[anchor=west, font=\small\bfseries] at (-1.20,-2.09)
  {Inside phase 3};
\draw[dashed, draw=black!50] (iterate.south) -- (6.70,-1.83);
\node[inner, fill=green!4] (hyp) at (0.30,-3.18)
  {$S$ parallel agents\\distinct hypotheses};
\node[inner, fill=green!4] (attempt) at (4.95,-3.18)
  {$I$ attempts per agent};
\node[inner, fill=green!4] (review) at (9.60,-3.18)
  {Fit, evaluate; meta-agent review};
\draw[flow] (hyp.east) -- (attempt.west);
\draw[flow] (attempt.east) -- (review.west);
\draw[flow, rounded corners=4pt]
  (review.south) -- ++(0,-0.94) -| (hyp.south);
\node[font=\scriptsize, fill=white, inner sep=2pt]
  at (4.95,-4.10) {repeat for $M$ meta-iterations};
\end{tikzpicture}
\caption{The four phases of \automodel{}.
In phase 3, parallel agents are spawned to test different structural hypotheses through sequential attempts, while a coordinating agent compares results and redirects the next meta-iteration.
The score and final check are defined for each application during setup.}
\label{fig:automodel-workflow}
\end{figure}
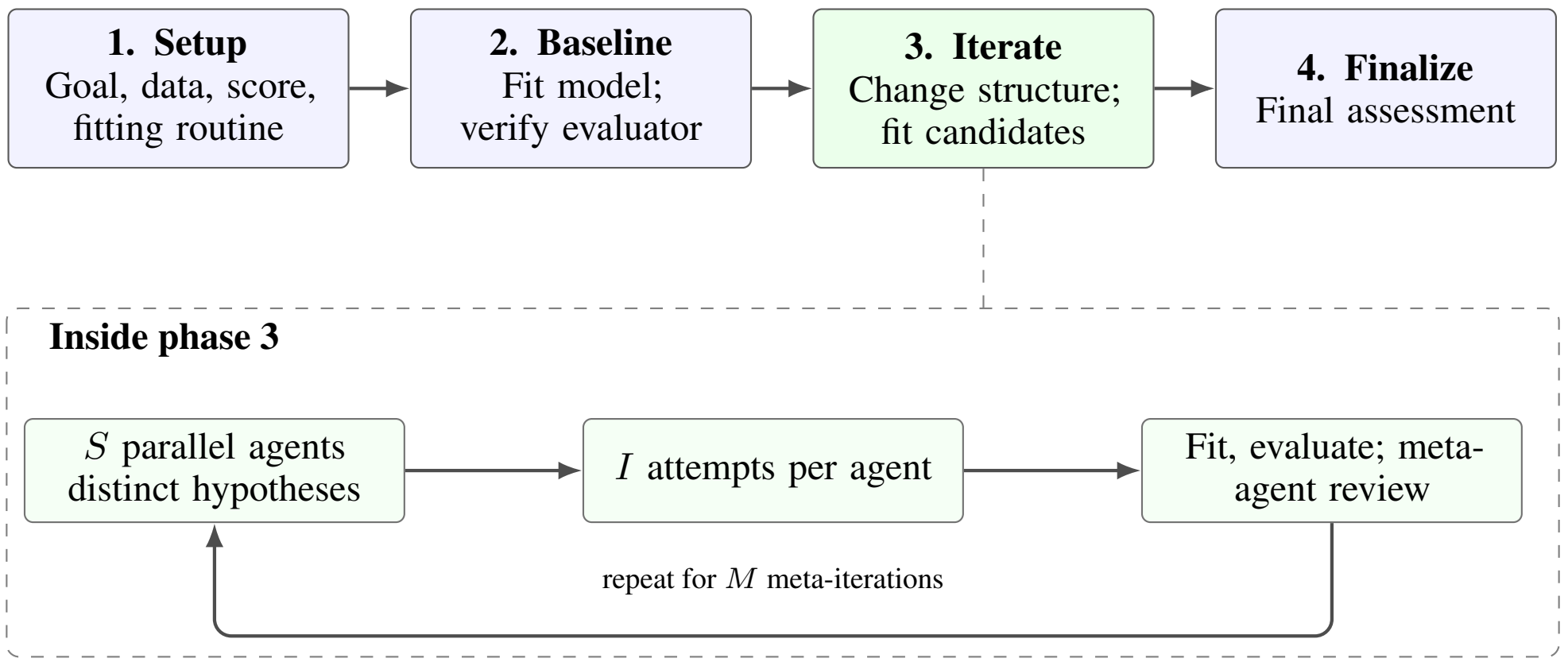

\automodel{} organizes model discovery into four sequential phases: \emph{setup}, \emph{baseline}, \emph{iterate}, and \emph{finalize} (Fig.~\ref{fig:automodel-workflow}).
Each phase has a separate recipe and produces contextual records for the next phase or to resume a previous run.

\paragraph{Phase 1: Setup.}
The researcher establishes the intended model with the agent, its inputs and outputs, admissible properties, fitting objective, selection score, and resource budget.
They implement or confirm the parameter-fitting routine (if any) that will be applied to a baseline model, and to new structural change attempts that will follow.
They also decide how candidate performance will be measured and what evidence will constitute an external final check.
The skill requires validation data or cross-validation for unpenalized error metrics such as root mean square error, whereas a complexity-penalized criterion (e.g. Akaike's information criterion~\cite{stoica2004model} or minimum description length~\cite{rissanen1978modeling})
may support selection on the fitted data alone.
An external final check is still needed, as penalizing model complexity does not account for selecting the best of many candidate structures.
Runtime and memory considerations
are assessed
so that the cost of evaluating candidates is known before the search.
Setup also checks that the environment can spawn sub-agents and execute the required experiments.
The agent records the decisions made during this phase in a \texttt{CONTEXT.md} file; this record is designed to keep track of choices and progress, and also to enable a later agent session to resume the work without relying on the conversation history. Progress is also tracked in a \texttt{CHECKLIST.md} file.

\paragraph{Phase 2: Baseline.}
This phase establishes a reference result and checks that candidate implementations can be compared through a working evaluator before structural search begins.
In particular, the agent fits the parameters of an existing model when one is supplied, and otherwise implements and fits a simple parameterized baseline.
A sub-agent then runs the complete evaluation pipeline, inspects whether the outputs are plausible, and records runtime and memory occupation, which will be used in Phase 3. This also verifies delegated file access and code execution.

\paragraph{Phase 3: Iterate.}
This phase implements the main search process, which consists of a nested loop where the agent acts as orchestrator, and sub-agents search in parallel.
Namely, the agent is tasked with spawning $S$ sub-agents and assigning them non-overlapping search directions.
The sub-agents are instructed to iteratively: (i) change the structure of the model according to their search direction, (ii) fit any free parameters, and (iii) record the respective results.
Once the sub-agents have completed $I$ \emph{iterations}, the orchestrating agent finalizes a \emph{meta-iteration} by reviewing the sub-agents' results and providing them with updated search directions, for them to start iterating again. 
It discards incomplete runs and independently re-evaluates shortlisted candidates using the common protocol before selecting a winner.
Once $M$ meta-iterations are completed, the agent coordinates with the user to decide whether to continue the search or move on to Phase 4.
Records of this phase are stored in structured folders containing the sub-agents' outputs, as well as in \texttt{CONTEXT.md} as summaries.

The parameters $S$ (number of parallel sub-agents), $I$ (number of sub-agent iterations), and $M$ (number of meta-iterations) are decided at the beginning of this phase in consultation with the user, using the time and memory occupation estimated on the baseline model in Phase 2.

\paragraph{Phase 4: Finalization.}
Once the user is content with the model(s) found, they are
assessed using the external check defined in Phase 1.
This may be an untouched test set, an independent dataset, and/or checks of physical properties beyond the fitted range.
If the result motivates further search, the same check should not be reused as independent evidence.
What is frozen before that assessment depends on the scientific question: some applications fix both structure and fitted coefficients, while others fix a shared structural mechanism but refit permitted case-specific parameters. The final model, performance, acceptance criteria, and any limitations are added to the experiment record in \texttt{CONTEXT.md}.

\vspace{3mm}

The following sections present five case studies in which different researchers applied \automodel{} to distinct scientific modeling problems.
Each study explains how the shared workflow was adapted to its data, model representation, objectives, and evaluation protocol, and reports the resulting model and its assessment.

\section{Case study 1: Astrophysics} 
\label{sec:astro}

\paragraph{Problem statement} A galaxy's stellar mass $M_\star$ is the total mass of
all the stars it contains, varying by roughly four orders of magnitude across the observed galaxy population. The \emph{stellar mass function} (SMF), written $\phi(M_\star)$, describes how common galaxies of each mass are: the number of galaxies per unit
volume of space, per factor-of-ten interval (``dex'') in stellar mass. It is one of the most basic
summary statistics in the study of how galaxies form, and provides a powerful constraint on any simulation or theory of galaxy
formation. Because it is measured by counting galaxies in mass
bins in a sky survey, the raw measurement is a histogram of counts, and is typically summarised with an analytic fitting formula. This is historically chosen
by eye, the literature standard being the Schechter function~\cite{schechter1976} with extensions proposed by e.g.
Saunders et al.~\cite{saunders1990} and Bernardi et al.~\cite{bernardi2013}.

\Ac{sr} provides a framework within which to search automatically for superior functional forms, overcoming the limited scope and potential confirmation bias of visual human fitting. Ford et al.~\cite{ford2026} applied Exhaustive Symbolic Regression (ESR)~\cite{bartlett2024}
--- which generates and scores \emph{all} functions up to a fixed complexity, and so is guaranteed to find both the optimal function and a complete ranking --- to the Sloan Digital Sky Survey (SDSS) survey measurement of the SMF by Bernardi et al.~\cite{bernardi2013}, using $x\equiv M_\star/10^{9}M_\odot$ as the
fitting variable. As the fit quality criterion, they adopted the Minimum Description Length
(MDL) principle, in the form developed for symbolic regression by
Bartlett et al.~\cite{bartlett2024}: the preferred function is the one that communicates the
data in the fewest nats (natural-logarithm units of information). For a function
$f(x;\theta)$ fitted to binned counts under a Poisson likelihood, the description length is
\begin{equation}
\mathrm{DL} \;=\; \underbrace{-\ln\mathcal{L}(\hat\theta)}_{\text{NLL}}
\;+\; \underbrace{\Big(\!-\tfrac{k}{2}\ln 3+\!\sum_j\big[\tfrac12\ln I_{jj}+\ln|\hat\theta_j|\big]\Big)}_{C_\theta\;\text{(parameters)}}
\;+\; \underbrace{n\ln c + \textstyle\sum\ln|\text{integers}|}_{C_f\;\text{(function)}},
\label{eq:dl}
\end{equation}
where $k$ is the number of fitted parameters, $I_{jj}$ the diagonal observed Fisher information, $n$
the number of nodes in the expression tree, and $c$ the number of distinct operators used.
DL approximates the negative log Bayesian evidence, rewarding goodness of fit (low negative log-likelihood NLL) while penalising in a principled way the addition of parameters ($C_\theta$) and operators ($C_f$). This makes it a robust single metric combining accuracy with complexity. A fitting function with lower DL compresses the data more efficiently.

Ref. \cite{ford2026} identifies functions, for example
$\phi = (\alpha + e^{\beta x^{\gamma}})^{\delta}$, that decisively beat the Schechter function, but did not reach the DL of the more complex Bernardi and Saunders forms. Being exhaustive, ESR cannot reach the complexities of the literature functions without enumerating every function within that complexity range, which is computationally prohibitive (realistically, ESR reaches complexity $\sim\!10$, while the Saunders and Bernardi functions contain $\gtrsim\!20$ operators). This motivates the use of a guided non-exhaustive search to attempt to bridge this gap, in this case \automodel.

\paragraph{Method} \automodel{} was used with Claude Code (Opus 4.8) with the following domain-specific choices on top of the baseline workflow.

\textit{(i) Objective and data organization.} The structure-selection score is the
description length of Eq.~\eqref{eq:dl}, while free parameters are fitted by maximizing the Poisson likelihood; its $C_\theta+C_f$ penalty is already an evidence-based
regulariser. We therefore dropped the baseline train/validation/test split and fitted on the
full binned data, as ESR does. We promoted generalisability in two additional ways. First, the
survey provides two versions of the same measurement, derived from two different ways of
inferring a galaxy's brightness and hence stellar mass from galaxy images (``S\'ersic'' and ``cmodel''); we treated
these as two datasets and required a candidate to win on \emph{both}. Second, following \cite{ford2026}, we imposed ``physicality'' checks on the functions to ensure that they describe physically sensible SMFs (see below).

\textit{(ii) Model and evaluation.}
A candidate ``model'' here is a one-line function string in ESR's operator basis ($+,-,\times,\div,\mathrm{pow},\exp,\log,\mathrm{inv}$). The shared train-and-evaluate harness is a single calibrated evaluator that fits the free parameters by Poisson likelihood (robust multi-start Nelder--Mead plus
basin-hopping) and returns the two complexity terms of Eq.~\eqref{eq:dl}. It reproduces the
DLs quoted in~\cite{bartlett2024} to $<\!0.01$ nat, so every candidate the agents generate is
directly comparable with the published numbers.

\textit{(iii) Search.} The agents were seeded with the already-evaluated ESR
and literature functions --- the current DL Pareto front --- as both targets and scaffolds.
We ran $M\!\times\!S\!\times\!I = 3\times4\times4$ (three meta-iterations of four parallel
agents, four sequential structural edits each)
and scoring every candidate on both datasets. In
practice, the agents reliably proposed families while the meta-agent performed the
definitive fits in controlled batches. Finally, all leading forms were re-optimized with a multi-method global search (massive multistart, differential evolution, dual annealing, basin-hopping), confirming that their margins reflect genuine structure rather than an
under-converged fit. The literature and ESR functions they are compared against are quoted
at their published fits (re-fitting the Bernardi function to the S\'ersic dataset with our evaluator lowers its DL by 14 nats, which would raise all S\'ersic $\Delta$DL values in Table~\ref{tab:dl} by the same amount).

\paragraph{Results}
The search returned seven functions that improve on the best exhaustive-search result; we
label them A1--A7 in order of increasing complexity. One of them, A5, beats the Bernardi
function in description length on both datasets:
\begin{equation}
\phi_{\rm A5}(M_\star) \;=\; x^{\theta_4}\,
\Big(\theta_0 + \theta_1^{\,x^{\theta_2}} + \theta_5\,\theta_6^{\,x}\Big)^{\theta_3},
\qquad x\equiv M_\star/10^{9}M_\odot,
\label{eq:a5}
\end{equation}
which is the ESR-like single component $(\theta_0+\theta_1^{x^{\theta_2}})^{\theta_3}$
augmented by one extra additive term, capturing the high-mass behaviour, and a power-law
prefactor capturing the low-mass behaviour. A simpler five-parameter form,
A2 $= x^{\theta_4}(\theta_0+\theta_1^{x^{\theta_2}})^{\theta_3}$, also beats Bernardi on one
of the two datasets. Table~\ref{tab:dl} shows the results for these functions.

\begin{figure}[htb]
\centering
\includegraphics[width=0.92\linewidth]{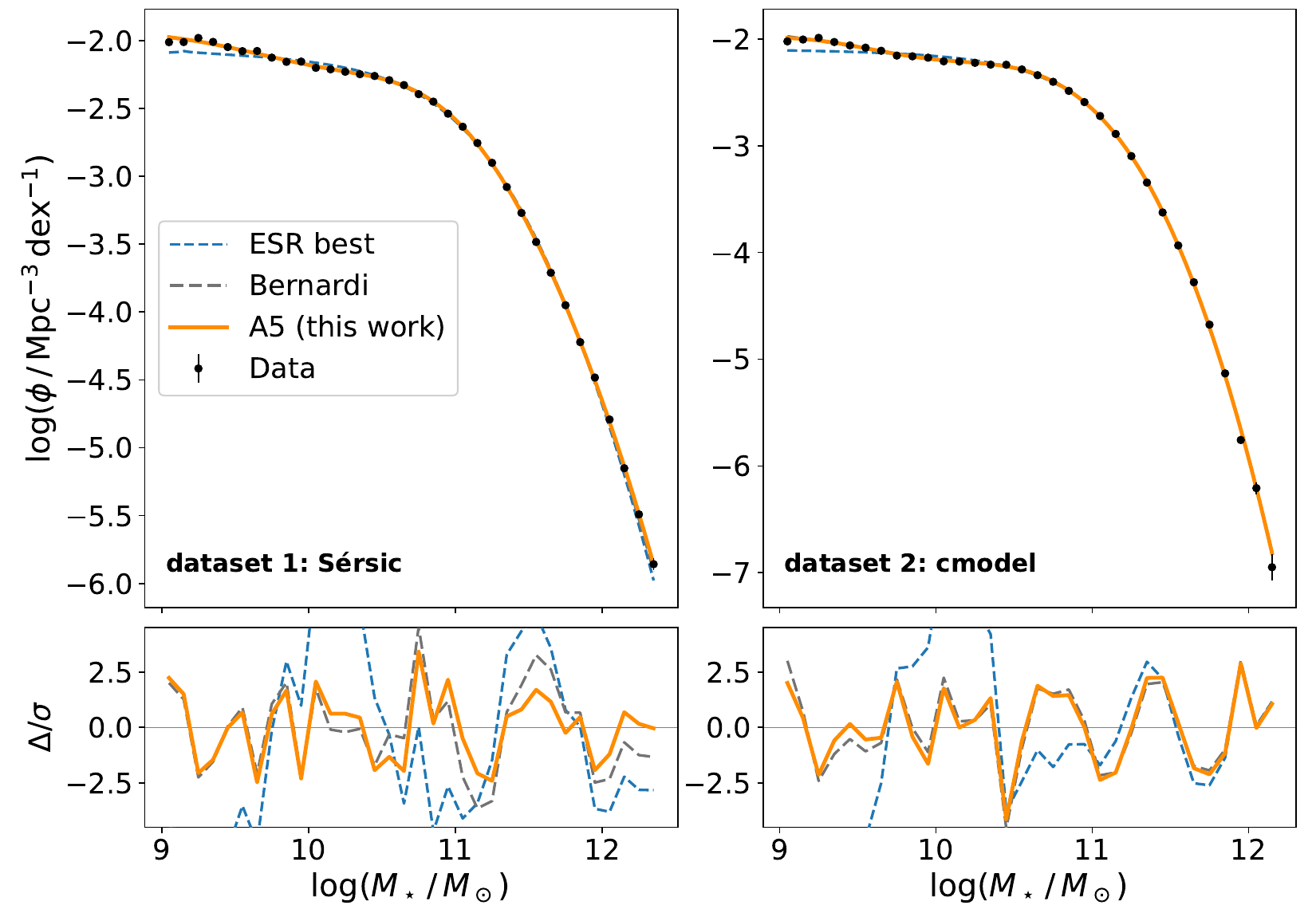}
\caption{The best ESR function, the Bernardi function, and the best function found by
\automodel{} (A5), overplotted on the binned SMF data. \emph{Top:} the number
density $\phi$ against stellar mass, both logarithmic, with the measurements as
points. \emph{Bottom:} residuals in units of the Poisson measurement uncertainty. A5 and Bernardi both
track the data to within about $2\sigma$, as expected for a high-quality fit, while the complexity-limited ESR function is visibly worse.}
\label{fig:funcs}
\end{figure}

A5 improves the description length relative to Bernardi by $36.5$ and $15.7$ nats on the two
datasets respectively, using 19 operators against Bernardi's 23. At the same time it improves the fit accuracy, as measured by a higher likelihood,
by replacing the Bernardi two-component structure with a generalised single component. Figure~\ref{fig:funcs} shows the functions overlaid on the binned data, together
with the residuals in units of the Poisson measurement uncertainty. A5 and Bernardi are visually
indistinguishable from the data and from each other, both scattering within about
$2\sigma$ of the data uncertainties; the best ESR function (i.e. the best function at complexity $\le$10) by contrast is visibly inaccurate, running low at the smallest masses and drifting to $3$--$4\sigma$ excursions through the intermediate range.

\begin{figure}[htb]
\centering
\includegraphics[width=0.92\linewidth]{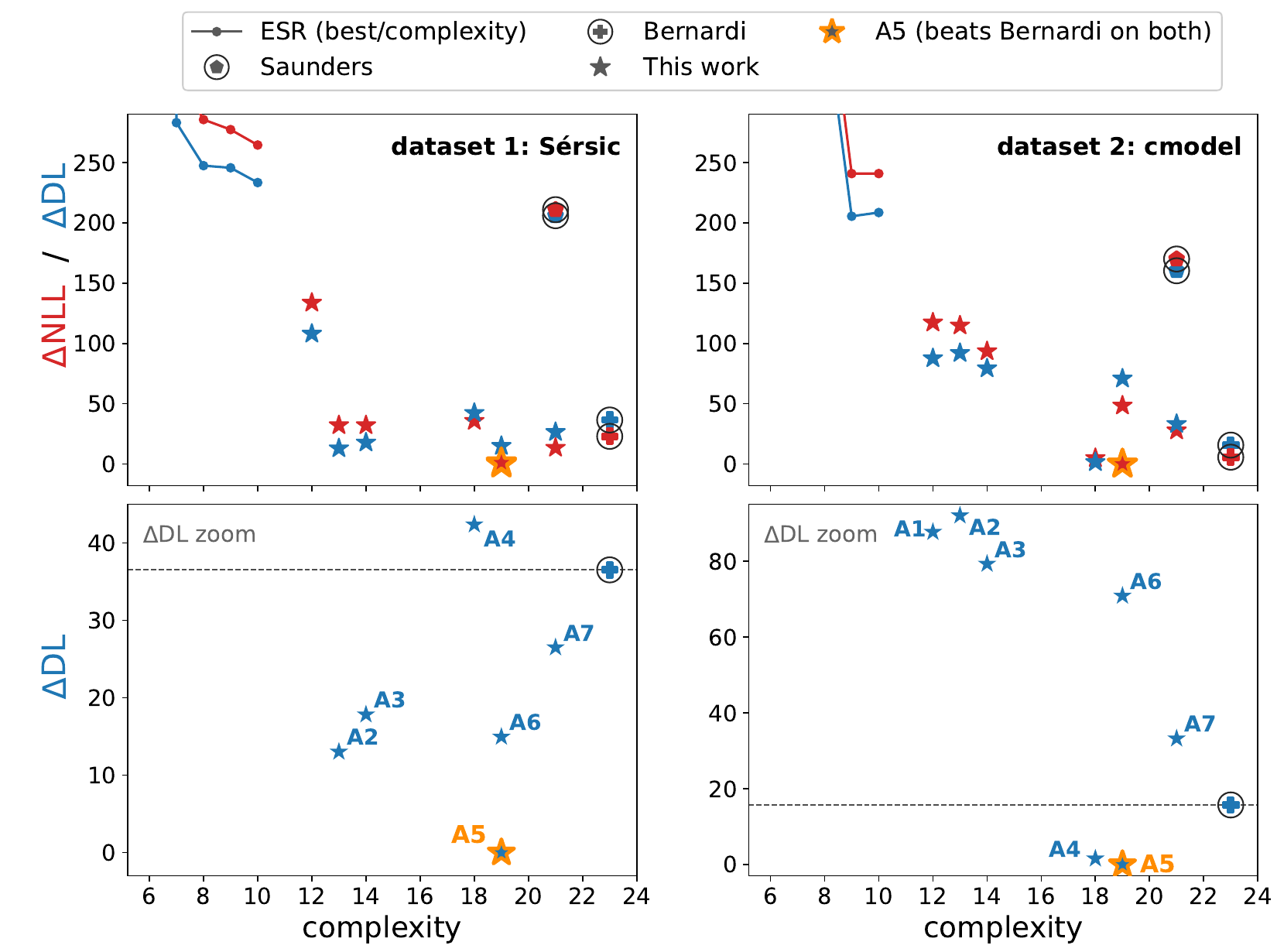}
\caption{Pareto fronts of $\Delta$DL (blue) and $\Delta$NLL (red) against complexity,
relative to the best value in each panel, for the two versions of the stellar mass function
dataset. Connected points are the best ESR function at each complexity, circled markers are
the Saunders and Bernardi literature forms and stars are the functions found by \automodel{},
labelled A1--A7 by increasing complexity. The lower
panels zoom in on the $\Delta$DL competition, the dashed horizontal line marking Bernardi's
value. The gold-rimmed A5 function beats Bernardi by description length on both datasets.
(Markers can coincide: two of the functions are at complexity 19, and A5's
$\Delta$DL and $\Delta$NLL markers lie exactly on top of one another on dataset 2.)}
\label{fig:pareto}
\end{figure}

Figure~\ref{fig:pareto} places the result on the Pareto front. The upper panels show the negative log-likelihood (red) and DL of each function, relative to the overall best (A5). The ESR functions lie to the upper left; i.e., relatively poor quality but simple. They are beaten in both likelihood and DL by the Saunders and Bernardi functions (circled symbols). The stars show the functions discovered by \automodel{}, which are significantly simpler than the literature functions while typically improving on both their likelihood and DL. The lower panel zooms in on the region around the lowest DL, showing that several of the \automodel{}-discovered functions beat Bernardi.

Beyond describing the data efficiently, a fitting formula for a number density must also be physically
admissible: it must stay positive (a negative number of galaxies is meaningless), it should
decrease as mass increases (reflecting the basic observational fact that there are more low-mass galaxies than massive ones), and its integral---the total mass contained in all stars per unit volume---must converge to a value below the total amount of baryonic mass available cosmologically. We verified all three for A5: it can safely be extrapolated beyond the measured range, and integrates to about $5\%$ of the cosmologically available mass budget. It is therefore the empirically best SMF derived to date.

\begin{table}[htb]
\centering
\small
\setlength{\tabcolsep}{5pt}
\caption{Goodness-of-fit statistics (description length DL and negative log-likelihood NLL) for the stellar mass function, defined relative to the Bernardi function (so negative numbers indicate superiority to Bernardi), for the two independent versions of the SMF dataset. $C$ is the number of operators in the expression (``complexity'' as defined by ESR), which is independent of the dataset except for the best ESR, which is different between the two datasets. A2 and A5 are the
\automodel{} functions discussed in the text; the best ESR function and the two literature forms are shown for reference.}
\label{tab:dl}
\begin{tabular}{l c r r c r r}
\toprule
 & & \multicolumn{2}{c}{dataset 1 (S\'ersic)} & & \multicolumn{2}{c}{dataset 2 (cmodel)}\\
\cmidrule(lr){3-4}\cmidrule(lr){6-7}
Function & $C$ & $\Delta$DL & $\Delta$NLL & & $\Delta$DL & $\Delta$NLL \\
\midrule
Best ESR                & 10/9 & $+196.8$ & $+241.5$ & & $+189.7$ & $+235.1$ \\
Saunders                & 21   & $+169.4$ & $+187.9$ & & $+144.6$ & $+164.3$ \\
Bernardi                & 23   & $0$      & $0$      & & $0$      & $0$      \\
A2 (this work)          & 13   & $-23.5$  & $+9.2$   & & $+76.3$  & $+109.1$ \\
\textbf{A5 (this work)} & \textbf{19} & $\mathbf{-36.5}$ & $\mathbf{-21.9}$ & & $\mathbf{-15.7}$ & $\mathbf{-5.7}$ \\
\bottomrule
\end{tabular}
\end{table}

\section{Case study 2: Cardiovascular modeling} 
\label{sec:use-case-2}

\paragraph{Problem statement}

Patient-specific cardiovascular modeling integrates clinical measurements into physiological models to reproduce patients' cardiovascular function.
Here, CircAdapt is considered, a computationally efficient closed-loop model of the heart and circulation that combines phenomenological descriptions of myocardial mechanics, cardiac chamber interaction, valves, and the systemic and pulmonary circulations \cite{Arts2005,Lumens2009,Walmsley2015}.
CircAdapt has been used extensively to simulate heart failure and electrical conduction disorders, where abnormal electrical activation patterns alter cardiac pump function.

Patient-specific use of such a model normally relies on calibrating its physiological parameters to clinical measurements. 
However, residual discrepancies between simulated and measured quantities may reflect limitations in the mathematical relationships that define the model itself.
Therefore, \automodel{} was used to test whether an agent-guided search could identify a physiologically interpretable modification of the CircAdapt equations.
The search was performed using data from multiple patients, requiring any newly introduced structural mechanism to be shared between patients while retaining independent calibration of the existing CircAdapt parameters.

\paragraph{Method}
\automodel{} was used with OpenAI Codex (GPT-5.6) with the following domain-specific choices on top of the baseline workflow.

\textit{(i) Objective and data organization.}
Cardiac magnetic resonance measurements from four patients with heart failure were considered, comprising six chamber-volume measurements per patient: left- and right-ventricular end-diastolic and end-systolic volumes and left- and right-atrial end-diastolic volumes.
For each patient, model fit was quantified across these six measurements using the mean absolute percentage error (MAPE),
\begin{equation}
    \mathrm{MAPE}
    =
    \frac{100}{6}
    \sum_{j=1}^{6}
    \left|
        \frac{V_j^{\mathrm{sim}}-V_j^{\mathrm{data}}}
             {V_j^{\mathrm{data}}}
    \right|.
    \label{eq:mape}
\end{equation}
Three patients were used for model discovery (``development patients''), while the fourth was held out for testing.

\textit{(ii) Model and evaluation.}
A candidate model was defined as a modification of the CircAdapt model equations.
The experiments used a C++ implementation of CircAdapt provided by Christoph M. Augustin and based on the formulation described in \cite{Augustin2021}.
To distinguish structural improvements from ordinary parameter fitting, the same 14 physiological and geometrical parameters were independently calibrated for each patient, whereas any parameter introduced by a structural modification was shared across the development patients.

Baseline and candidate models were evaluated using the same patient-specific inputs, parameter bounds, convergence criteria, post-processing, and differential-evolution calibration.
Simulations that failed to converge or produced invalid chamber volumes were rejected.

Each structural modification included an identity value that exactly recovered the original CircAdapt formulation, providing a direct consistency check for the implementation and evaluation pipeline.

In a preliminary exploration on Patient~1, \automodel{} proposed a modification to the right-ventricular reference area that passed the identity check. A subsequent audit of the implementation showed that the new coefficient was algebraically equivalent to rescaling an existing CircAdapt parameter. We therefore included all five cardiac reference areas among the 14 parameters calibrated separately for each patient in the subsequent multi-patient searches.

\textit{(iii) Search.}
Following the single-patient exploration, the multi-patient search proceeded in two stages.
The first multi-patient search used $M\times S\times I = 1\times3\times1$, corresponding to one meta-iteration, three parallel hypothesis agents, and one structural proposal per agent.

The agents were not restricted to modifying a predefined equation or physiological subsystem.
The resulting candidate models targeted distinct mechanisms in myocardial activation, right-ventricular pressure--geometry coupling, and atrial--pericardial interaction.

Parameter fitting distinguished between shared structural coefficients and patient-specific CircAdapt parameters.
Shared structural coefficients were screened via grid search.
For each grid value, the 14 ordinary CircAdapt parameters were recalibrated independently for each patient. Candidate structures were compared using the mean of the patient-level MAPEs across the three development patients.

For the held-out patient, the best-found structural formulation and its shared coefficient were kept fixed, while the same 14 patient-specific CircAdapt parameters were recalibrated.
This provided a verification of the model selected on the development cohort without using the held-out patient for structural discovery.

After completion of the held-out evaluation, a second and broader development-only search was run to test whether the selected mechanism remained competitive against a wider range of structural alternatives.
This second search used $M \times S \times I = 2 \times 4 \times 2$,
corresponding to two meta-iterations, four parallel hypothesis streams per meta-iteration, and two sequential structural attempts per stream.
All 16 proposed equation modifications were implemented and evaluated using only the original three development patients.
The search spanned multiple components of CircAdapt, including myocardial mechanics, chamber interaction, valves, circulation, and pericardial mechanics.
The held-out patient was not used at any stage of this broader search.

\paragraph{Results}

The first multi-patient search produced three structurally distinct candidates modifying myocardial activation, right-ventricular pressure--geometry coupling, and atrial--pericardial interaction.
The strongest improvement was obtained by modifying the myocardial activation model.

In the original CircAdapt model, myocardial contraction depends on the stretch of an internal contractile element within the muscle cells~\cite{Augustin2021}. This dependence is described by the activation factor $C_s$, which depends on $L_{\mathrm{norm}}$, the normalized extension of this internal contractile element, according to
\begin{equation}
    C_s = \tanh\left( 6.84 L_{\mathrm{norm}}^2 \right).
    \label{eq:circadapt_original}
\end{equation}
The modification found with \automodel{} extends this relation by also accounting for the instantaneous deformation of the myocardial tissue:
\begin{equation}
    C_s^{*} = \tanh\left( 6.84 L_{\mathrm{norm}}^2 \exp(g_L E_f) \right),
    \label{eq:circadapt_automodel}
\end{equation}
where $g_L$ is a shared coefficient and $E_f$ measures this tissue-level deformation.
The modified equation therefore combines information about the internal contractile state with the mechanical deformation experienced by the tissue.
The shared coefficient was screened over $g_L \in \{0,1,2,3,4\}$, with $g_L=3$ giving the lowest mean development error among the tested values.

\begin{figure}[ht]
    \centering
    \includegraphics[width=\linewidth]{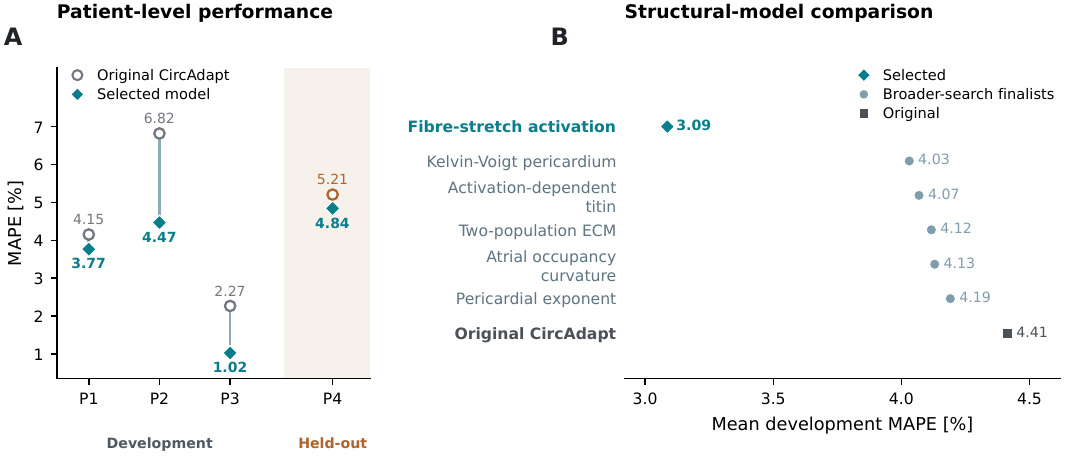}
    \caption{
    Performance of the structural model identified by \automodel.
    \textbf{Panel A}: MAPE of the calibrated original CircAdapt formulation and the selected model. P1--P3 denote the development patients and P4 the held-out patient.
    \textbf{Panel B}: mean development MAPE of the selected model, the calibrated original formulation, and the five finalists retained from the subsequent search of 16 additional structural mechanisms. Lower MAPE indicates better agreement with the patient-specific measurements.
    }
    \label{fig:circadapt_results}
\end{figure}

After full patient-specific recalibration, the selected modification reduced MAPE in all three development patients compared to the original CircAdapt formulation: from $4.15\%$ to $3.77\%$ in Patient 1, from $6.82\%$ to $4.47\%$ in Patient 2, and from $2.27\%$ to $1.02\%$ in Patient 3 (Fig.~\ref{fig:circadapt_results}A). 

For the held-out patient 4, MAPE decreased from $5.21\%$ to $4.84\%$.
This smaller aggregate improvement was not uniform across individual chamber volumes, and is therefore interpreted as a modest held-out advantage rather than evidence of population-level generalisation.

In the broader second search, no newly tested modification outperformed the previously selected fibre-stretch activation model on the development patients.
The best newly discovered alternative reached a mean development MAPE of $4.03\%$, while the five finalists from this broader search ranged from $4.03\%$ to $4.19\%$, compared with $3.09\%$ for the selected model.
Thus, the activation modification identified in the first multi-patient search remained the best-performing structure among the broader set of tested mechanisms (Fig.~\ref{fig:circadapt_results}B).

\section{Case study 3: Precision Agriculture}\label{sec:use-case-3} 

\paragraph{Problem statement}

Optimizing Nitrogen (N) fertilization in precision agriculture requires understanding site-specific crop responses to input factors.
This challenge is hindered by traditional approaches that rely either on predefined parametric forms~\cite{bullock94,CF}, which may be too simplistic to capture diverse response patterns across a field, or on opaque machine learning models (e.g., neural networks), which limit interpretability~\cite{MZ}.
To account for spatial variability in field characteristics like soil properties and terrain, fields are delineated into management zones (MZs)~\cite{managementNrate}.
MZs are distinct subregions within a field that exhibit similar agronomic characteristics affecting crop yield.
They are designed to help farmers treat these subregions separately, tailoring practices like N fertilizer application to local conditions to improve crop productivity while minimizing environmental impact~\cite{managementNrate}.

Mathematically, the true yield function for a field site at coordinates $(\texttt{lat}, \texttt{lon})$ is represented by a complex multivariate system denoted as $y_{(\texttt{lat}, \texttt{lon})} = f(\mathbf{X}_{(\texttt{lat}, \texttt{lon})})$. 
However, N-rate optimization does not focus on the absolute yield values, but rather on the functional form between the expected yield at $(\texttt{lat}, \texttt{lon})$ and the N rate.
This functional form can be interpreted as the shape of the curve that relates the N rate and yield response, which defines the N-response curve.
Since the absolute values are not necessary for optimization, any vertical shifts in N-response curves are eliminated to obtain the relative N-response curve.
Hence, N-rate optimization isolates the functional relationship between the estimated relative crop yield, denoted as $\texttt{ry}_{(\texttt{lat}, \texttt{lon})}$, and all admissible fertilizer rates $\mathbf{x}^{Nr}$, fixed between $0$ and $150$ pounds per acre (lbs/acre) for winter wheat.
As such, the relative N-response curve is denoted as $\mathbf{x}^{Nr}$ vs. $\texttt{ry}_{(\texttt{lat}, \texttt{lon})}$.

Given a global set of pre-computed N-response curves for all field sites, denoted as $\mathbf{R}$, the specific subset of curves associated with the $z$-th MZ is defined as $\mathbf{R}_z$.
Building on the assumption that all sites within an MZ share an underlying functional form~\cite{MZ}, the objective of this problem is to identify this shared structure for each MZ $z$, expressed as a symbolic functional skeleton $\hat{\mathbf{e}}_z(\mathbf{x}^{Nr})$. 
A symbolic skeleton acts as an abstract mathematical template, capturing the structural form of the expression using constant placeholders rather than specific numerical values.
Once a generalized skeleton is discovered for a given MZ, these placeholders are optimized to minimize the RMSE for each site within that zone, yielding a parametric site-specific N-response function.

\paragraph{Method} 

\begin{figure}[!t]
    \centering
    \begin{subfigure}{0.2\columnwidth}
        \centering
        \includegraphics[width=\linewidth]{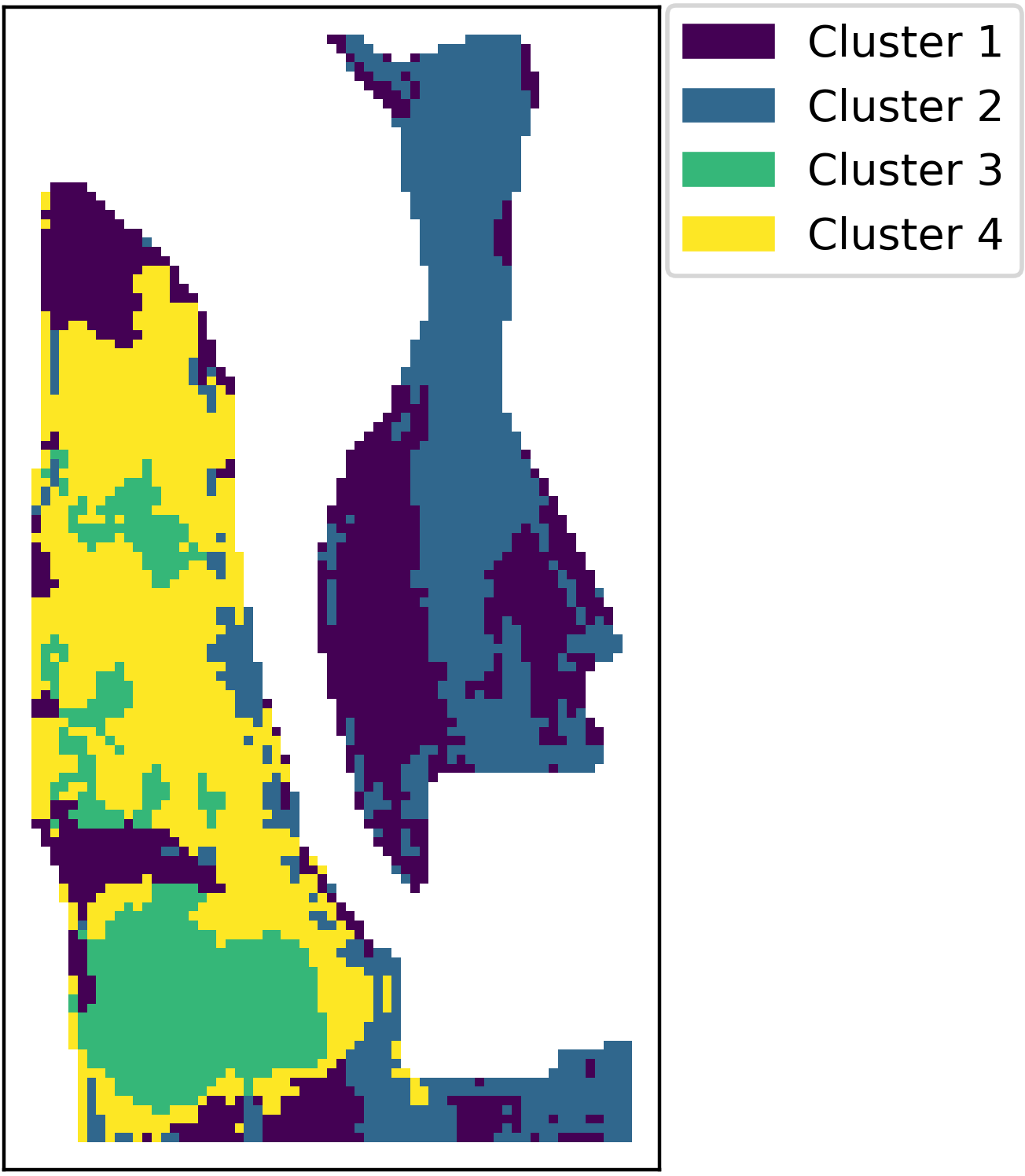} 
    \vspace{-3ex}
        \caption{}
        \label{fig:image_a1}
    \end{subfigure}
    \begin{subfigure}{0.45\columnwidth}
        \centering
        \includegraphics[width=\columnwidth]{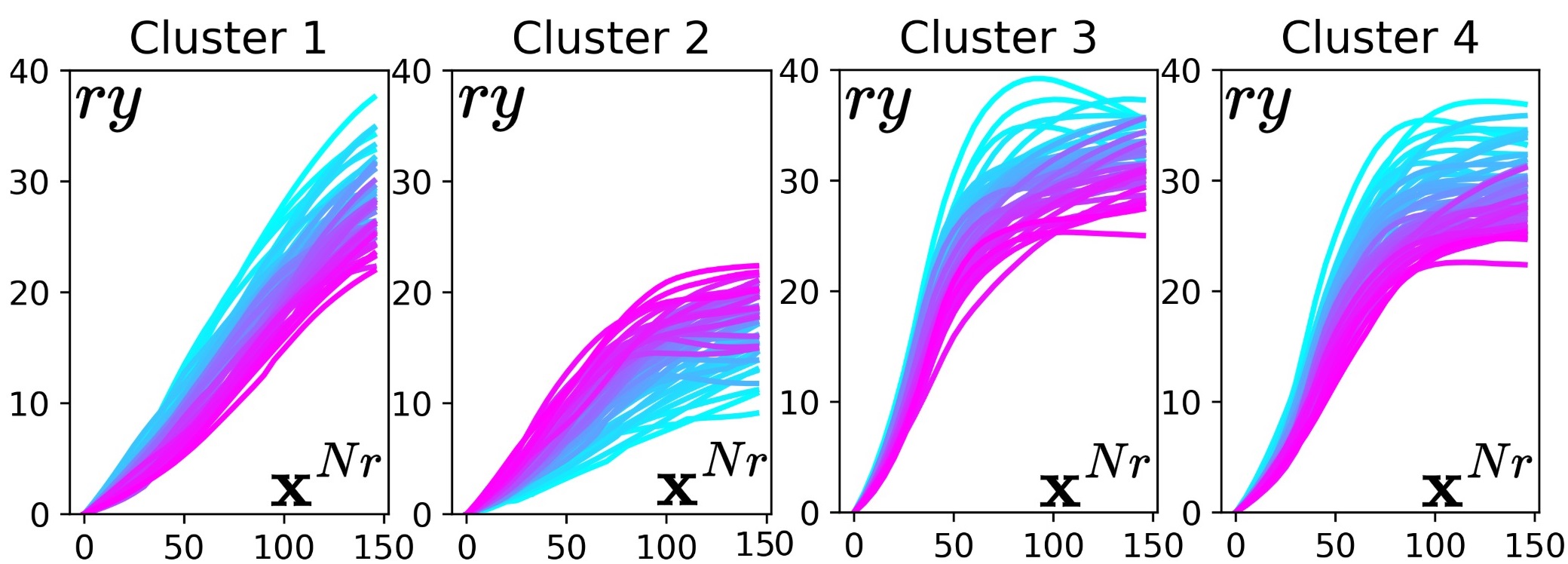} 
    \vspace{0ex}
        \caption{}
        \label{fig:image_b1}
    \end{subfigure}
    \caption{Results obtained for Field A. \textbf{(a)} Delineated management zones and \textbf{(b)} Aligned approximated N-response curves for each MZ (N-rate $\mathbf{x}^{Nr}$ vs. relative yield $\texttt{ry}$)~\cite{MZ}.}
    \label{fig:clusteredA}
\end{figure}

\begin{figure}[!t]
    \centering
    \begin{subfigure}{0.2\columnwidth}
        \centering
        \includegraphics[width=\linewidth]{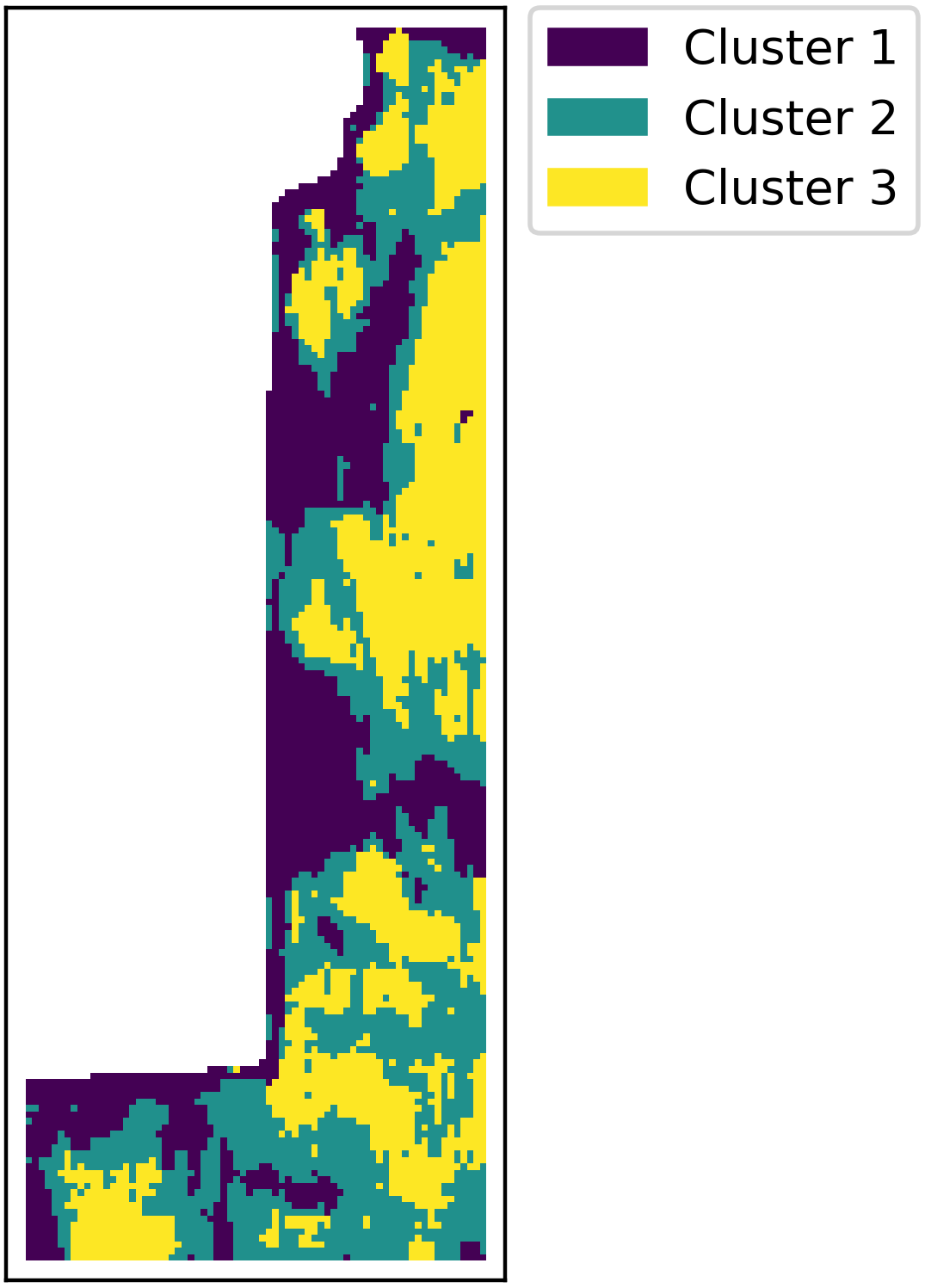} 
    \vspace{-3ex}
        \caption{}
        \label{fig:image_a}
    \end{subfigure}
    \begin{subfigure}{0.45\columnwidth}
        \centering
        \includegraphics[width=\columnwidth]{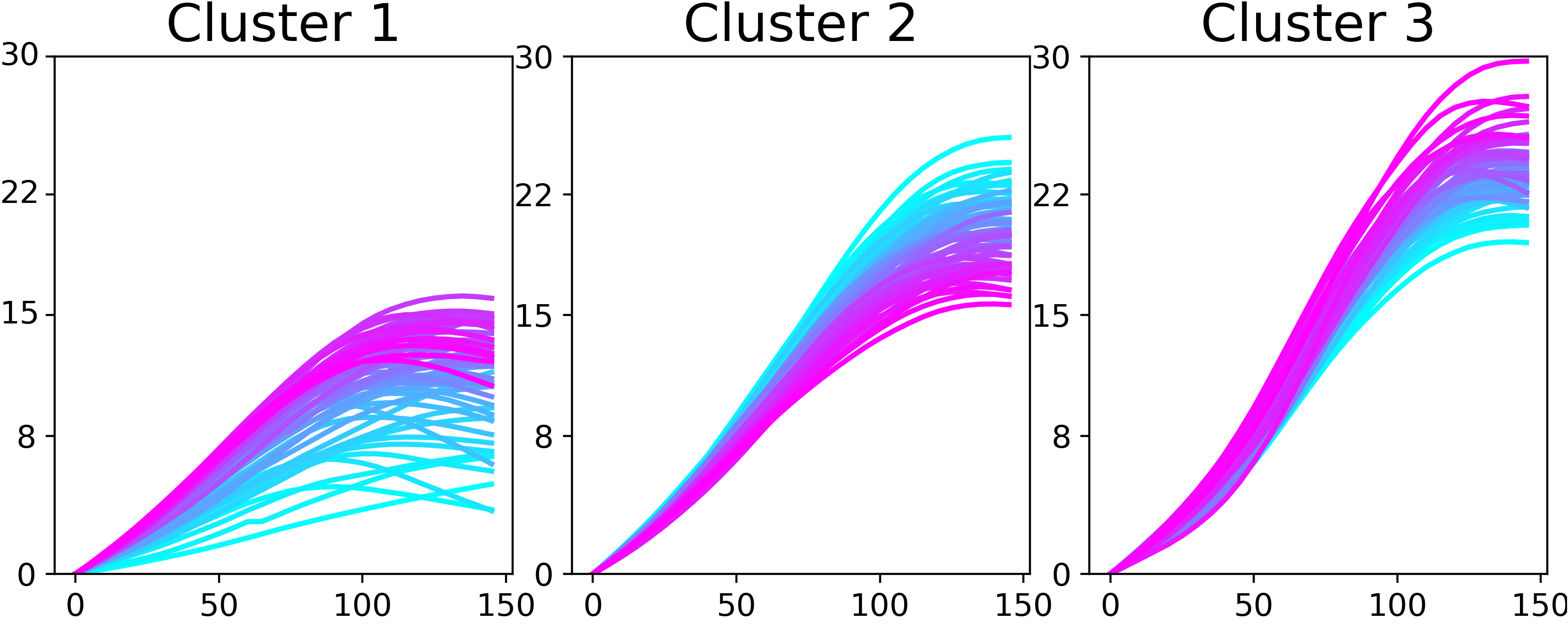} 
    \vspace{0ex}
        \caption{}
        \label{fig:image_b}
    \end{subfigure}
    \caption{Results obtained for Field B. \textbf{(a)} Delineated management zones and \textbf{(b)} Aligned approximated N-response curves for each MZ (N-rate $\mathbf{x}^{Nr}$ vs. relative yield $\texttt{ry}$)~\cite{MZ}.}
    \label{fig:clusteredB}
\end{figure}

\automodel{} was used with Claude Code (Opus 5) with the following domain-specific choices:

\textit{(i) Objective and data organization.}
The data utilized here were derived from two winter wheat dryland fields located in Montana, USA, hereafter referred to as Field A and Field B.
The field is spatially discretized into a grid, where each point or ``site" corresponds to a $10 \times 10$ m area.
Comprehensive details regarding the dataset, the formulation of the early-yield prediction problem, its 2D–3D convolutional neural network (CNN) prediction model, the N-response curve generation process, and the MZ clustering method are documented in~\cite{MZ}.
For this specific problem, Fields A and B have already been delineated into four and three distinct MZs, respectively, as shown in Figs.~\ref{fig:clusteredA} and~\ref{fig:clusteredB}, and the N-response curves $\mathbf{R}_z$ have been pre-computed for each $z$-th MZ in both fields using a CNN.
Under this partitioning, Field A consists of four MZs ($z \in \{1, 2, 3, 4\}$) containing 305, 289, 467, and 639 individual sites, respectively. 
Similarly, Field B comprises three MZs ($z \in \{1, 2, 3\}$) containing 1,024, 1,566, and 1,623 individual sites, respectively.

For this application, we deviate from the standard \automodel{} procedure by selecting models using in-sample error and performing finalization refitting on the same full dataset, rather than evaluating on a held-out set as the skill normally requires for unpenalized accuracy metrics.
Because this constitutes a descriptive and explanatory task rather than a traditional predictive modeling problem requiring train-test data splitting, no data splits were considered.
Note that the objective is to discover a single representative skeleton $\hat{\mathbf{e}}_z(\mathbf{x}^{Nr})$ that effectively characterizes the shared agronomic response of an entire cluster. 
Consequently, the skeleton is derived using the complete collection of curves in $\mathbf{R}_z$ and is then fitted to all sites within that specific cluster to yield localized parameterizations, rendering data holdouts unnecessary for this stage of functional discovery.

\textit{(ii) Model and evaluation.}
The search space was defined as the set of closed-form, univariate, and differentiable mathematical skeleton expressions $\hat{\mathbf{e}}_z(\mathbf{x}^{Nr})$, whose free parameters are the constant placeholders $c_1, \dots, c_k$.
Each candidate structure encapsulates three components: the skeleton expression, a data-driven routine that produces an initial guess for the placeholders from an individual curve, and per-placeholder box bounds.
The automated search navigates this space by iteratively mutating the skeletons' mathematical formulation while leaving the underlying evaluation protocol unaltered.
For a given MZ, a skeleton is fitted independently to each of the $\vert{}\mathbf{R}_z\vert{}$ curves.

For a given field site in the $z$-th MZ at coordinates $(\texttt{lat}, \texttt{lon})$, the parameter fitting process is formulated to minimize the sum of squared residuals, $ \sum_{j} \big(\hat{\mathbf{e}}_z(\mathbf{x}^{Nr}_j; \mathbf{c}_{(\texttt{lat}, \texttt{lon})}) - \texttt{ry}_{(\texttt{lat}, \texttt{lon}),\, j}  \big)^2$, across all $j$ sampled points of the site's curve for the parameter vector $\mathbf{c} = [c_1, \dots, c_k]^T \in \mathbb{R}^k$.
Here, $\hat{\mathbf{e}}_z(\mathbf{x}^{Nr}_j; \mathbf{c}_{(\texttt{lat}, \texttt{lon})})$ is the instantiated function; that is, $\hat{\mathbf{e}}_z(\mathbf{x}^{Nr}_j)$ with its constant placeholders replaced by the specific parameter values in vector $\mathbf{c}_{(\texttt{lat}, \texttt{lon})}$.
Note that each target curve $\texttt{ry}_{(\texttt{lat}, \texttt{lon})}$ was previously standardized via curve-wise $z$-score normalization (zero mean and unit variance across its sampled points).
Because raw response curves across different sites vary substantially in vertical scale and amplitude, standardizing each curve individually places all sites on a common dimensionless scale.
This prevents high-variance curves from dominating cluster-level aggregated error metrics during model selection.
It is worth pointing out that the agents were not given a fitting routine; instead, they designed and implemented one themselves as part of \automodel{}'s \textbf{Phase 1}, and then reused it for the entire study.
Hence, since the generated functional forms are frequently highly non-convex with respect to their parameters $\mathbf{c}$, the agents employed the Trust-Region Reflective (TRF) algorithm, a bounded multi-start nonlinear least-squares optimization strategy.

Consistent with the descriptive nature of the task, the evaluator reports in-sample fit quality across all sites within an MZ. 
The performance is quantified by reporting the mean error $\bar{r}$ over all $\vert{}\mathbf{R}_z\vert{}$ field sites belonging to the $z$-th MZ: $\bar{r} = \frac{1}{\vert{}\mathbf{R}_z\vert{}} \sum_{{(\texttt{lat}, \texttt{lon})} \in \, z\text{-th MZ}} {\left\vert{} \hat{\mathbf{r}}_{(\texttt{lat}, \texttt{lon})} - \texttt{ry}_{(\texttt{lat}, \texttt{lon})} \right\vert}$,
where $\hat{\mathbf{r}}_{(\texttt{lat}, \texttt{lon})}$ represents the corresponding fitted N-response curve vector for that site; i.e., the instantiated skeleton $\hat{\mathbf{e}}_z(\mathbf{x}^{Nr}; \mathbf{c}^*_{(\texttt{lat}, \texttt{lon})})$ using the already optimized parameter values $\mathbf{c}^*_{(\texttt{lat}, \texttt{lon})}$ for that location.
Because the objective is to characterize the underlying functional shape of observed agronomic responses rather than to forecast unmeasured locations, no held-out data split is utilized, and no claims of generalization to unseen sites are made.
For the purposes of this evaluation, the pre-computed response curves are treated as the ground-truth baseline; i.e., $\bar{r}$ is used strictly as a goodness-of-fit metric to demonstrate which functional skeleton most accurately recovers their shape.
While evaluating fidelity to these baseline curves inherently relies on the assumption that the CNN has accurately captured predictive agronomic responses, the objective of this stage remains descriptive: assessing the structural flexibility of the candidate skeletons in approximating these functional forms.

\textit{(iii) Search.}
The \automodel{} skill was executed independently for each field.
In Phase 1, the objective and data preprocessing steps were verified, followed by a shape-profile analysis of the target curves, which revealed important domain characteristics.
For instance, 81.2\% of sites in Field B's MZ~1 showed yield declines at higher N rates, indicating that standard monotonically saturating functional families were structurally incapable of representing that zone.

Phase 2 established initial benchmarks using a sigmoidal baseline, $c_1\tanh(c_2 x + c_3) + c_4$. 
For Field B, the best-performing skeletons, previously discovered for Field A, were also evaluated as transfer baselines. 
Their poor performance confirmed that optimal functional forms do not directly transfer across distinct fields, highlighting the need for dedicated discovery per field.

Phase 3 executed the core discovery loop, structured as a multi-agent search parameterized by $M$ meta-iterations $\times$ $S$ parallel agents $\times$ $I$ sequential inner attempts per agent. 
Field A was configured with $M=2, S=3, I=3$.
Field B was initialized with $M=1, S=3, I=3$; however, runtime interruptions terminated two agents prematurely (after zero and one iteration, respectively), leaving the third agent to complete all three iterations. 
Hence, the orchestrator formulated and evaluated several additional candidates directly within that single meta-iteration.
Across both fields, each agent explored a distinct, non-overlapping functional space, such as explicitly non-monotonic peaked expressions, higher-order Padé rational kernels, or composite warped-argument forms, by proposing incremental structural modifications to the candidate mathematical expressions. 
Between meta-iterations, an orchestrator synthesized the performance of all proposed modifications, separating helpful structural changes from refuted hypotheses to guide subsequent iterations.

Human-in-the-loop consultation occurred at three decision points: (1) defining the search loop hyperparameters ($M, S, I$), (2) reviewing cross-agent candidate rankings before finalizing, and (3) arbitrating a complexity trade-off on Field B. 
For Field B, extending a seven-parameter skeleton to an eight-parameter variant yielded a 17--41\% per-zone error reduction, exceeding the pre-registered threshold of $>5\%$ improvement per added parameter, at the cost of increased parameter ill-conditioning. 
Upon reviewing the trade-off, the user selected the eight-parameter formulation.

Finally, Phase 4 performed final fits for the selected skeletons across all field sites using expanded multi-start optimization runs. 
An independent verification process confirmed the exact numerical reproducibility of the stored equations and residual errors. 
A coefficient-identifiability audit was also conducted, establishing, as a stated limitation, that while the final eight-parameter curves are highly accurate and reproducible, their individual site-level parameter constants exhibit weak identifiability and should not be interpreted independently across sites.

\paragraph{Results}
Tables~\ref{tab:FieldA} and~\ref{tab:FieldB} report the skeletons discovered by \automodel{} for each MZ of Fields A and B, together with those produced by the neural SR-based method SeTGAP~\cite{NeuroSRPA} and by the two classical parametric N-response models, quadratic-plateau~\cite{bullock94} and exponential~\cite{CF}. 
For every method, every MZ, and every field site, the placeholders of the corresponding skeleton were fitted with the same optimizer, described in the previous section, 
to minimize the distance to that site's N-response curve, and we report the resulting mean error over all sites of the zone.

\begin{table}[t]
\centering
\caption{Field A: Symbolic skeletons per MZ and mean error $\bar{r}$ over all sites of the zone.}
\vspace{1ex}
\label{tab:FieldA}
\footnotesize
\setlength{\tabcolsep}{4pt}
\begin{tabular}{c l l r}
\toprule
MZ & Method & Skeleton $\mathbf{e}(x)$ & $\bar{r}$ \\
\midrule
\multirow{4}{*}{1}
 & \automodel{} & $c_1\tanh\!\big(c_2 x + c_3 - c_4\sqrt{(x-c_5)^2+1}\big)+c_6$ & 0.0043 \\ 
 & SeTGAP & $c_1\tanh(c_2 x + c_3)+c_4$ & 0.0113 \\
 & Quadratic-plateau & $c_1 + c_2\min(c_3,x)^2$ & 0.5411 \\
 & Exponential & $c_1\big(1-e^{\,c_2+c_3 x}\big)+c_4$ & 0.0636 \\
\midrule
\multirow{4}{*}{2}
 & \automodel{} & $(c_1+c_2 x)\big((c_3+c_4 x)^2+1\big)^{-c_5}+c_6$ & 0.0067 \\ 
 & SeTGAP & $c_1\tanh(c_2 x + c_3)+c_4$ & 0.0185 \\
 & Quadratic-plateau & $c_1 + c_2\min(c_3,x)^2$ & 0.4107 \\
 & Exponential & $c_1\big(1-e^{\,c_2+c_3 x}\big)+c_4$ & 0.0902 \\
\midrule
\multirow{4}{*}{3}
 & \automodel{} & $(c_1+c_2 x)\big((c_3+c_4 x)^2+1\big)^{-c_5}+c_6$ & 0.0077 \\ 
 & SeTGAP & $c_1 x + c_2\cos(c_3 x + c_4)+c_5$ & 0.0553 \\
 & Quadratic-plateau & $c_1 + c_2\min(c_3,x)^2$ & 0.2286 \\
 & Exponential & $c_1\big(1-e^{\,c_2+c_3 x}\big)+c_4$ & 0.1234 \\
\midrule
\multirow{4}{*}{4}
 & \automodel{} & $(c_1+c_2 x)\big((c_3+c_4 x)^2+1\big)^{-c_5}+c_6$ & 0.0075 \\ 
 & SeTGAP & $\exp\!\big(c_1\sin(c_2 x + c_3)\big)$ & 0.0251 \\
 & Quadratic-plateau & $c_1 + c_2\min(c_3,x)^2$ & 0.3226 \\
 & Exponential & $c_1\big(1-e^{\,c_2+c_3 x}\big)+c_4$ & 0.1469 \\
\bottomrule
\end{tabular}
\end{table}

\begin{table}[t]
\centering
\caption{Field B: Symbolic skeletons per MZ and mean error $\bar{r}$ over all sites of the zone. }
\vspace{1ex}
\label{tab:FieldB}
\footnotesize
\setlength{\tabcolsep}{4pt}
\begin{tabular}{c l l r}
\toprule
MZ & Method & Skeleton $\mathbf{e}(x)$ & $\bar{r}$ \\
\midrule
\multirow{4}{*}{1}
 & \automodel{} & $(c_1+c_2 x)\big((c_3 x + c_4 - c_5\sqrt{(x-c_6)^2+1})^2+1\big)^{-c_8}+c_7$ & 0.0021 \\ 
 & SeTGAP & $c_1 x^4 + c_2 x^3 + c_3 x^2 + c_4 x + c_5$ & 0.0139 \\
 & Quadratic-plateau & $c_1 + c_2\min(c_3,x)^2$ & 0.4044 \\
 & Exponential & $c_1\big(1-e^{\,c_2+c_3 x}\big)+c_4$ & 0.1401 \\
\midrule
\multirow{4}{*}{2}
 & \automodel{} & $(c_1+c_2 x)\big((c_3 x + c_4 - c_5\sqrt{(x-c_6)^2+1})^2+1\big)^{-c_8}+c_7$ & 0.0016\\ 
 & SeTGAP & $c_1\big(c_2\sin(c_3 x)+c_4\big)^{-1}+c_5$ & 0.0204 \\
 & Quadratic-plateau & $c_1 + c_2\min(c_3,x)^2$ & 0.4933 \\
 & Exponential & $c_1\big(1-e^{\,c_2+c_3 x}\big)+c_4$ & 0.0990 \\
\midrule
\multirow{4}{*}{3}
 & \automodel{} & $(c_1+c_2 x)\big((c_3 x + c_4 - c_5\sqrt{(x-c_6)^2+1})^2+1\big)^{-c_8}+c_7$ &  0.0020\\ 
 & SeTGAP & $c_1\tanh(c_2 x + c_3)+c_4$ & 0.0214 \\
 & Quadratic-plateau & $c_1 + c_2\min(c_3,x)^2$ & 0.5281 \\
 & Exponential & $c_1\big(1-e^{\,c_2+c_3 x}\big)+c_4$ & 0.1221 \\
\bottomrule
\end{tabular}
\end{table}

The two classical agronomic models are not competitive in either field.
Averaged over zones, \automodel{}-discovered formulations are $57.4\times$ more accurate than quadratic-plateau and $16.2\times$ more accurate than the exponential model on Field A, and $248.9\times$ and $63.1\times$ more accurate, respectively, on Field B.
Notably, the agents observed that the failure of the quadratic-plateau model is structural rather than numerical.
That is, both classical families are monotone and saturating by construction, whereas the agents observed that 38\% of the sites of Field A's MZ 4 and 81.2\% of the sites of Field B's MZ 1 exhibit a yield decline at high N rates.
A monotone-saturating form cannot express an interior maximum, so for those sites it must either flatten through the peak or misplace the plateau, and the fitted error is dominated by that structural mismatch.
Against SeTGAP, \automodel{} performs better on average across both fields, being $4.2\times$ more accurate on Field A and $9.7\times$ more accurate on Field B, though no statistical significance is implied, as formal tests have not been performed. 
In all cases, \automodel{} requires using more complex mathematical expressions with a greater number of fitting coefficients (e.g., up to 8 coefficients on Field B compared to SeTGAP's maximum of 5). 
While this added complexity provides the flexibility needed to capture the response curve shapes and reduce fit error, it potentially challenges the interpretability of the resulting equations compared to simpler forms.


\section{Case study 4: Automotive tire modeling}\label{sec:use-case-4} 

\paragraph{Problem statement}
In vehicle dynamics and automotive engineering, accurate modeling of tire dynamics is fundamental to developing reliable digital twins and control algorithms that improve vehicle performance, safety, and handling. Indeed, tires constitute the only interface through which forces are transmitted between the vehicle and the road, and their behavior largely determines the overall dynamic response of ground vehicles.

Modeling pneumatic tires, however, is inherently challenging because of the complex physical phenomena involved, including viscoelastic material behavior, transient contact dynamics, load transfer, temperature effects, and nonlinear friction mechanisms. As a result, the conventional approach in vehicle dynamics relies on empirical or semi-empirical constitutive laws that express tire forces as functions of longitudinal and lateral slip. In particular, the most widely used model is Pacejka's Magic Formula, which postulates a nonlinear relationship of the form~\cite{Pacejka}
\begin{align}\label{eq:yGenPacejka}
    y = D\sin[C\arctan(Bx-E[Bx-\arctan(Bx)])], \quad x \in \mathbb{R},
\end{align}
where $y \in \mathbb{R}$ is a generic normalized tire output (normalized longitudinal force, lateral force, or aligning moment), $x \in \mathbb{R}$ is the independent variable (longitudinal or lateral slip), and the macro-parameters $B,C,D,E \in \mathbb{R}$ may depend on additional micro-parameters such as normal load and tire temperature. For instance, Eq.~\eqref{eq:yGenPacejka} may be applied to model the lateral tire force $F_y \in \mathbb{R}$ as a function of the lateral slip $\sigma_y \in \mathbb{R}$:
\begin{align}\label{eq:yGenPacejkaFy}
    F_y(\sigma_y) = F_z\mu_y(\sigma) = F_zD\sin[C\arctan(B\sigma_y-E[B\sigma_y-\arctan(B\sigma_y)])], \quad \sigma_y \in \mathbb{R}.
\end{align}
Alternative expressions to~\eqref{eq:yGenPacejkaFy} can now be discovered using AI-based equation discovery methods. Indeed, since~\eqref{eq:yGenPacejkaFy} is a purely empirical model rather than a physics-derived constitutive law, it offers no intrinsic advantage over alternative empirical models identified from data.

\paragraph{Method}
The \automodel{} skill was used with Codex (GPT-5) to explore alternative analytical model structures. 
The workflow was adapted to the present tire-modelling problem through the following domain-specific choices. 

\textit{(i) Objective and data splitting.}
The objective was to discover an analytical model for the normalized lateral tire force $\mu_y(\sigma_y)$.
The measurements were organized into five vertical-load bins with representative loads \begin{equation} F_z = \left\{ 929.685,\, 1934.457,\, 2934.066,\, 3886.111,\, 4890.660 \right\} \ \mathrm{N}. \label{eq:representative_vertical_loads} \end{equation} 
Each load bin was represented by $200$ sampled pairs $(\sigma,\mu_y)$. The three lowest-load bins were used jointly for parameter identification, resulting in $600$ training observations. The fourth bin was reserved exclusively for structural validation and model selection, while the fifth bin was retained as an untouched test set. Consequently, the fifth bin was not used for parameter identification, structural hypothesis generation, or candidate-model ranking. The primary model-selection metric was the root-mean-square error of the normalized force, \begin{equation} \mathrm{RMSE}_{\mu} = \sqrt{ \frac{1}{N} \sum_{i=1}^{N} \left( \widehat{\mu}_{y,i}-\mu_{y,i} \right)^2 }. \label{eq:normalised_rmse} \end{equation} The median absolute error, \begin{equation} \mathrm{MedAE}_{\mu} = \operatorname{median}_{i} \left| \widehat{\mu}_{y,i}-\mu_{y,i} \right|, \label{eq:normalised_median_error} \end{equation} and the coefficient of determination, \begin{equation} R^2 = 1- \frac{ \sum_{i=1}^{N} \left( \widehat{\mu}_{y,i}-\mu_{y,i} \right)^2 }{ \sum_{i=1}^{N} \left( \mu_{y,i}-\overline{\mu}_y \right)^2 }, \label{eq:normalised_r2} \end{equation} were used as complementary accuracy measures. Because $F_z$ is constant within each bin, the dimensional RMSE for a given bin can be recovered according to \begin{equation} \mathrm{RMSE}_{F_y} = F_z\,\mathrm{RMSE}_{\mu}. \label{eq:dimensional_rmse} \end{equation} 

\textit{(ii) Model and evaluation.}
A load-dependent Pacejka's Magic Formula~\eqref{eq:yGenPacejkaFy}, without horizontal and vertical shifts, was adopted as the baseline.
This choice ensured that the benchmark corresponded to the prescribed four-factor Pacejka structure rather than to an extended shifted Magic Formula. To improve numerical conditioning, the vertical load was represented by the centered variable \begin{equation} q(F_z) = \frac{F_z-F_{z,\mathrm{ref}}}{1000\ \mathrm{N}}, \label{eq:normalised_vertical_load} \end{equation} 
with $F_{z,\mathrm{ref}} = 2934$ N, and the Pacejka macro-parameters were assumed to vary smoothly with load according to \begin{subequations} \label{eq:pacejka_macroparameter_load_dependence} \begin{align} B(F_z) &= \exp\!\left(b_0+b_1q(F_z)\right),\\ C(F_z) &= c_0+c_1q(F_z),\\ D(F_z) &= d_0+d_1q(F_z),\\ E(F_z) &= e_0+e_1q(F_z) , \quad F_z \in \mathbb{R}_{\geq 0}. \end{align} \end{subequations} The exponential parametrization of $B$ guarantees a positive stiffness factor over the investigated load range. The remaining macro-parameters were assigned affine load dependencies to retain a compact and interpretable baseline. 

All eight secondary coefficients, \begin{equation} \boldsymbol{\theta}_{\mathrm{P}} = \begin{bmatrix} b_0 & b_1 & c_0 & c_1 & d_0 & d_1 & e_0 & e_1 \end{bmatrix}^{\mathrm{T}}, \label{eq:pacejka_secondary_parameters} \end{equation} were identified simultaneously using the three training bins.
Therefore, the baseline was not fitted independently at each load. Instead, a single set of secondary coefficients defined continuous functions of $F_z$ shared across the training conditions. 

\textit{(iii) Search.}
After calibration of the baseline, the GPT-5 reasoning model assisted in generating alternative analytical structures. The structural search was subject to the following constraints: \begin{enumerate} \item The resulting model had to remain explicit and analytical; \item The response had to remain bounded over the measured slip interval; \item The model had to retain a continuous dependence on the vertical load; \item Polynomial basis expansions were excluded; \item Candidate structures could employ natural functions such as $\tanh$, $\arctan$, $\exp$, and related smooth compositions; \item The final candidate had to represent a genuinely non-Pacejka structure rather than merely adding standard shifts to the Magic Formula. \end{enumerate} 

The search used $M \times S \times I = 3\times3\times5$, thus employing three parallel agents with different hypotheses and three meta-iterations; the explored candidate families included extended Pacejka formulations, hyperbolic-tangent saturation models, and inverse-tangent saturation models. Standard horizontal and vertical Pacejka shifts were included during the exploratory stage as consistency references. However, shifted or asymmetrically scaled Pacejka formulations were not considered novel model structures for the final non-Pacejka comparison. Among the genuinely alternative structures, the best validation performance was obtained with the load-dependent asymmetric hyperbolic-tangent model
\begin{equation}  \mu_y(\sigma_y) = S_{\mathrm{v}}(F_z) + D(F_z) \tanh\left( K(F_z)\sigma_{\mathrm{h}}(\sigma_y) \right)\left[ 1+ A(F_z) \tanh\left( \frac{\sigma_{\mathrm{h}}(\sigma_y)}{\sigma_0} \right) \right], \quad \sigma_y \in \mathbb{R}  \label{eq:natural_asymmetric_tanh} \end{equation} where \begin{equation} \sigma_{\mathrm{h}}(\sigma_y) \triangleq \sigma-S_{\mathrm{h}}(F_z). \label{eq:natural_effective_slip} \end{equation} In Eq.~\eqref{eq:natural_asymmetric_tanh}, the first hyperbolic-tangent term describes the transition from the approximately linear low-slip response to force saturation, the second hyperbolic-tangent term provides a smooth sign-dependent modulation, allowing the model to represent different response levels on the positive and negative slip branches, and the parameters $S_{\mathrm{h}}$ and $S_{\mathrm{v}}$ account for systematic horizontal and vertical offsets without requiring separate models for the two slip directions. The load dependencies were defined as \begin{subequations} \label{eq:natural_macroparameter_load_dependence} \begin{align} D(F_z) &= d_0+d_1q(F_z),\\ K(F_z) &= \exp\!\left(k_0+k_1q(F_z)\right),\\ S_{\mathrm{h}}(F_z) &= s_{\mathrm{h},0}+s_{\mathrm{h},1}q(F_z),\\ S_{\mathrm{v}}(F_z) &= s_{\mathrm{v},0}+s_{\mathrm{v},1}q(F_z),\\ A(F_z) &= a_0+a_1q(F_z), \quad F_z \in \mathbb{R}_{\geq 0}. \end{align} \end{subequations} In particular, the exponential representations of $K$ ensures a positive transition-rate parameter. The complete secondary-parameter vector of the natural model was \begin{equation} \boldsymbol{\theta}_{\mathrm{N}} = \begin{bmatrix} d_0 & d_1 & k_0 & k_1 & s_{\mathrm{h},0} & s_{\mathrm{h},1} & s_{\mathrm{v},0} & s_{\mathrm{v},1} & a_0 & a_1  \end{bmatrix}^{\mathrm{T}}. \label{eq:natural_secondary_parameters} \end{equation} 

For each candidate structure, the secondary coefficients governing both the slip response and its load dependence were identified jointly from bins 1--3, and parameter estimation was performed using bounded nonlinear least-squares optimization in MATLAB\textsuperscript{\textregistered}. Multiple initial parameter vectors were used for every structure to reduce sensitivity to local minima. The same training observations, parameter bounds, convergence tolerances, number of initializations, and performance metrics were used for all candidate structures, and the load-dependent secondary coefficients were shared across the three training bins. Moreover, no model coefficient was fitted independently for a single training bin. This constraint required each candidate to explain both the slip dependence and the systematic variation with vertical load through one continuous analytical formulation. After fitting on bins 1--3, each candidate and its identified parameters were evaluated on bin 4 without recalibration. Candidate structures were ranked primarily according to validation RMSE, while the median absolute error and $R^2$ were inspected to ensure that the ranking did not result from a small number of unusually large residuals. Parameter vectors reaching prescribed optimization bounds were also recorded, since such behavior may indicate parameter correlation or limited practical identifiability. The asymmetric hyperbolic-tangent structure in Eq.~\eqref{eq:natural_asymmetric_tanh} was selected as the best non-Pacejka model because it provided the lowest validation RMSE among the genuinely alternative analytical structures considered. Once this structure and its load-dependent secondary coefficients had been selected, both were frozen before the fifth bin was evaluated. 

For the held-out fifth bin, neither the analytical structure nor any of its secondary coefficients was modified.
In contrast to, e.g., the patient-specific calibration procedures for cardiovascular modeling (Section~\ref{sec:use-case-2}), no parameters were recalibrated on the test bin, because the objective was to assess direct extrapolation of the complete load-dependent tire model to a previously unused vertical-load condition. The final test evaluation therefore consisted only of inserting $F_z=4891$ N into the previously identified macro-parameter functions and evaluating the resulting force curve at the supplied test values of $\sigma_y$. The test RMSE, median absolute error, and $R^2$ were computed only after completion of validation-based model selection. The GPT-5 reasoning model was used to assist structural hypothesis generation, analytical formulation, and implementation of the candidate models, whereas all secondary coefficients were obtained from explicit numerical optimization, and all training, validation, and test metrics were computed directly from the experimental data. The final model structure, its implementation, the identified coefficients, and the resulting predictions were inspected by the author.

\paragraph{Results} The model-discovery procedure selected the load-dependent asymmetric hyperbolic-tangent formulation~\eqref{eq:natural_effective_slip} as the most accurate non-Pacejka analytical structure among the candidates considered. On the three training bins, the natural model reduced the normalized-force RMSE from 0.059 for the Pacejka baseline to 0.052, corresponding to a reduction of approximately 12.0\%. The median absolute error decreased from 0.0367 to 0.0317, while the coefficient of determination increased from $R^2=0.987$ to $R^2=0.990$. The improvement was retained on the independent validation bin, for which the RMSE decreased from 0.0467 to 0.0384, an improvement of approximately 17.8\%, and the median absolute error decreased from 0.0331 to 0.0280. The corresponding validation coefficient of determination increased from 0.990 to 0.993. After model selection, the previously unused fifth load bin was evaluated once as the final test set. The natural model achieved a test RMSE of 0.0546, compared with 0.0583 for the Pacejka baseline, and reduced the median absolute error from 0.0518 to 0.0463. The test coefficient of determination consequently increased from 0.983 to 0.986. At the representative test load $F_z=4891$ N, these normalized RMSE values correspond to dimensional force errors of approximately 266.9 N for the natural model and 284.9 N for the Pacejka baseline. Although the relative improvement on the test set, approximately 6.3\%, was smaller than that observed during validation, the natural model consistently outperformed the baseline across the training, validation, and test partitions. The results indicate that the combination of load-dependent saturation, stiffness, horizontal and vertical offsets, and a smooth sign-dependent modulation captures systematic features of the measured lateral-force response that are not represented by the unshifted Pacejka baseline. Nevertheless, because some identified coefficients approached their prescribed optimization bounds, the individual macro-parameters should be interpreted primarily as components of the resulting predictive law rather than as independently identifiable physical quantities.

\begin{figure}
    \centering
    \includegraphics[width=1\linewidth]{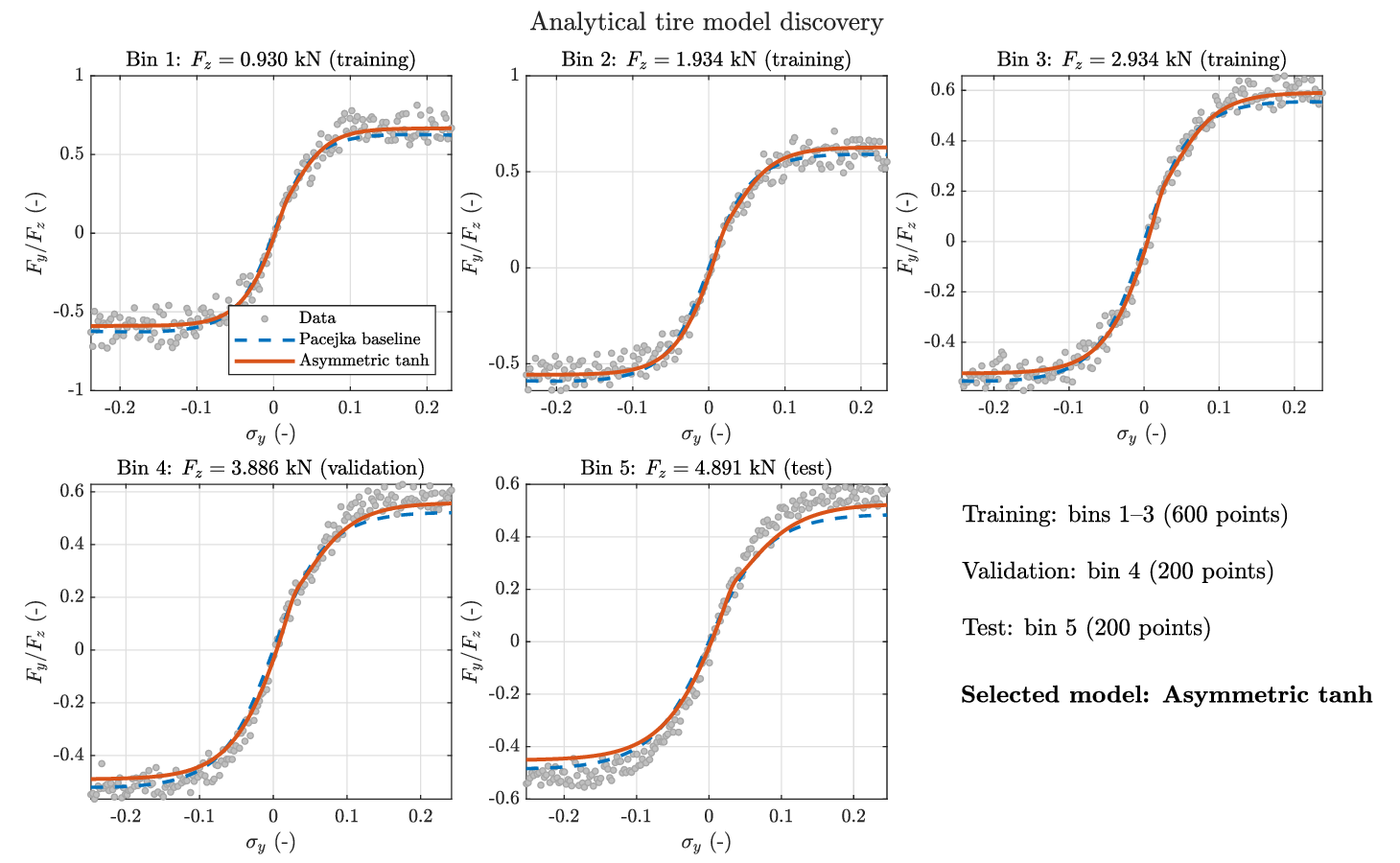}
    \caption{Comparison between the Pacejka Magic Formula baseline (dashed blue lines), the selected natural model (orange lines), and the experimental data.}
    \label{fig:TireModelComp}
\end{figure}

A comparison between the Pacejka's Magic Formula baseline and the selected model, with parameter values listed in Table~\ref{tab:TireParam}, and the experimental data is illustrated in Fig.~\ref{fig:TireModelComp}, whereas the relevant performance indicators are reported in Table~\ref{tab:performaceTire}.

\begin{table}[t] \centering \caption{Identified load-dependent parameters of the Pacejka baseline and the natural asymmetric hyperbolic-tangent model. The normalized load variable is $q=(F_z-F_{z,\mathrm{ref}})/(1000\,\mathrm{N})$, with $F_{z,\mathrm{ref}}=2934\,\mathrm{N}$. The coefficients multiplying $q$ represent sensitivities per unit load variation expressed in $\mathrm{kN}$.} 
\label{tab:TireParam}
\small \begin{tabular}{llllr} \toprule Model & Macro-parameter & Functional dependence & Secondary parameter & Value \\ \midrule Pacejka baseline & $B(F_z)$ & $\exp(b_0+b_1q)$ & $b_0$ ($-$) & $2.46$ \\ & & & $b_1$ ($\mathrm{kN}^{-1}$) & $-0.136$ \\[1mm] & $C(F_z)$ & $c_0+c_1q$ & $c_0$ ($-$) & $1.39$ \\ & & & $c_1$ ($\mathrm{kN}^{-1}$) & $0.00444$ \\[1mm] & $D(F_z)$ & $d_0+d_1q$ & $d_0$ ($-$) & $0.558$ \\ & & & $d_1$ ($\mathrm{kN}^{-1}$) & $-0.0318$ \\[1mm] & $E(F_z)$ & $e_0+e_1q$ & $e_0$ ($-$) & $0.409$ \\ & & & $e_1$ ($\mathrm{kN}^{-1}$) & $-0.0122$ \\ \midrule Natural model & $D(F_z)$ & $d_0+d_1q$ & $d_0$ ($-$) & $0.562$ \\ & & & $d_1$ ($\mathrm{kN}^{-1}$) & $-0.0296$ \\[1mm] & $K(F_z)$ & $\exp(k_0+k_1q)$ & $k_0$ ($-$) & $2.67$ \\ & & & $k_1$ ($\mathrm{kN}^{-1}$) & $-0.145$ \\[1mm] & $S_{\mathrm{h}}(F_z)$ & $s_{\mathrm{h},0}+s_{\mathrm{h},1}q$ & $s_{\mathrm{h},0}$ ($-$) & $0.0240$ \\ & & & $s_{\mathrm{h},1}$ ($\mathrm{kN}^{-1}$) & $0.00432$ \\[1mm] & $S_{\mathrm{v}}(F_z)$ & $s_{\mathrm{v},0}+s_{\mathrm{v},1}q$ & $s_{\mathrm{v},0}$ ($-$) & $0.200$ \\ & & & $s_{\mathrm{v},1}$ ($\mathrm{kN}^{-1}$) & $0.00461$ \\[1mm] & $A(F_z)$ & $a_0+a_1q$ & $a_0$ ($-$) & $-0.298$ \\ & & & $a_1$ ($\mathrm{kN}^{-1}$) & $-0.0241$ \\[1mm] & $\sigma_0$ & & ($-$) & $0.0116$ \\ \bottomrule \end{tabular} \end{table}

\begin{table}[t]
\centering \caption{Prediction performance of the Pacejka baseline and the natural asymmetric hyperbolic-tangent model. Model parameters are identified jointly using load bins 1--3. Bin 4 is used for validation, whereas bin 5 is retained as the final test set. RMSE and median absolute error are evaluated for the normalized lateral force $\mu_y=F_y/F_z$.} 
\label{tab:performaceTire} 
\begin{tabular}{llccc} \toprule Dataset & Model & RMSE & $\operatorname{median}|e|$ & $R^2$ \\ \midrule Training, bins 1--3 & Pacejka baseline & $0.0592$ & $0.0367$ & $0.9869$ \\ & Natural model & $\mathbf{0.0521}$ & $\mathbf{0.0317}$ & $\mathbf{0.9899}$ \\ \midrule Validation, bin 4 & Pacejka baseline & $0.0467$ & $0.0331$ & $0.9900$ \\ & Natural model & $\mathbf{0.0384}$ & $\mathbf{0.0280}$ & $\mathbf{0.9932}$ \\ \midrule Test, bin 5 & Pacejka baseline & $0.0583$ & $0.0517$ & $0.9835$ \\ & Natural model & $\mathbf{0.0546}$ & $\mathbf{0.0463}$ & $\mathbf{0.9855}$ \\ \bottomrule \end{tabular} \end{table}

\section{Case study 5: Traffic flow modeling}\label{sec:use-case-5} 
\label{sec:traffic-flow}

\paragraph{Problem statement}
Macroscopic traffic models describe the evolution of vehicle density
$\rho(x,t)$ along a road. In the Lighthill--Whitham--Richards (LWR) model,
density satisfies the scalar conservation law
\begin{equation}
    \partial_t \rho + \partial_x q(\rho) = 0,
    \label{eq:traffic-lwr}
\end{equation}
where the fundamental diagram $q(\rho)$ relates density to traffic flux and
the corresponding velocity is $v(\rho)=q(\rho)/\rho$.
Classical fundamental
diagrams provide compact and physically interpretable constitutive laws, but a
fixed analytical form can retain systematic bias when it is embedded in a
spatio-temporal traffic simulation.
Replacing the diagram entirely with a
flexible data-driven model may improve fit at the cost of interpretability and
physical structure.

The intermediate hypothesis is considered here, that a calibrated
classical diagram captures the principal traffic physics while its remaining
bias can be represented by a compact multiplicative correction:
\begin{equation}
    q_{\mathrm{corr}}(\rho)
    = q_{\mathrm{base}}(\rho;\boldsymbol{\theta})
      g(\rho;\boldsymbol{c}),
    \label{eq:traffic-correction}
\end{equation}
where $q_{\mathrm{base}}$ is one of five established diagrams---Greenshields,
the equilibrium Intelligent Driver Model (IDM), Weidmann, Triangular, or Del
Castillo---whose calibrated coefficients $\boldsymbol{\theta}$ remain fixed
during structural discovery; the correction $g$ is dimensionless, and
$g=1$ recovers the unmodified diagram. Restricting the selected corrections to
positive exponential multipliers preserves the units, sign, and zeros of the
baseline flux, while still allowing local and short-range spatial adjustments.

\paragraph{Method}
\automodel{} was used with OpenAI Codex (GPT-5.6 Sol) with the following domain-specific choices on top of the baseline workflow.
The implementation, configurations, and reproducibility materials are available in the
\href{https://github.com/cpml-au/automodel-traffic}{\texttt{automodel-traffic}
repository}.

\textit{(i) Objective and data organization.}
We used density, velocity, and flow measurements from the NGSIM Interstate~80
data set. After nondimensionalization, the data comprise 79 spatial rows and
180 time samples; 75 interior spatial positions enter the objective. The split
was chronological and fixed before model discovery: samples 0--63 were used to
fit correction coefficients, samples 64--107 for structural validation and
model selection, and samples 108--179 as a final held-out test interval. No
shuffling was used. The two component errors are relative $L_2$ errors,
\begin{equation}
    E_\rho = \frac{\lVert\widehat{\rho}-\rho\rVert_2}
                        {\lVert\rho\rVert_2},
    \qquad
    E_v = \frac{\lVert\widehat{v}-v\rVert_2}
                    {\lVert v\rVert_2},
\end{equation}
and candidate fit was summarized by
\begin{equation}
    E_{\mathrm{data}} = 50\left(E_\rho^2+E_v^2\right).
    \label{eq:traffic-data-error}
\end{equation}
Structure selection used validation
$E_{\mathrm{fitness}}=E_{\mathrm{data}}+0.01n$, where $n$ is the number of
nodes in the expression tree. Thus, an added mechanism had to offset its
complexity cost rather than merely produce a numerically negligible reduction
in residual error.

\textit{(ii) Model and evaluation.}
A candidate model was a multiplier $g$ applied to one of the five fixed
fundamental diagrams in Eq.~\eqref{eq:traffic-correction}. Each candidate was
evaluated by solving Eq.~\eqref{eq:traffic-lwr} with the same initial state,
measured boundary density, mesh, and time refinement as its baseline. The
finite-volume update used a Rusanov/Godunov flux implemented with discrete
exterior calculus operators. For nonlocal corrections, the wave-speed bound
included the off-diagonal terms of the flux Jacobian, rather than treating a
spatially coupled flux as pointwise.

The coefficient optimizer used deterministic bounded Powell searches with
explicit starts and evaluation budgets. Candidates were rejected if their
multipliers or simulations were non-finite, if they produced a non-positive
multiplier or negative velocity on the relevant physical density domain, or if
the corrected velocity violated the prescribed monotonicity check. Every
expression, parameter vector, train/validation score, feasibility result,
runtime, and memory measurement was retained.

\textit{(iii) Search.}
The search used $M\times S\times I=5\times3\times3$: five meta-iterations,
three parallel hypothesis streams per iteration, and three sequential
structural attempts per stream. Every proposed structure was fitted separately
to all five baseline diagrams, giving 225 baseline--candidate fits before final
refitting on the combined train and validation interval. Early iterations explored positive pointwise polynomial, anchored,
saturating, and rational corrections. Later iterations introduced short-range
convolutions and Hodge-star features defined on the traffic mesh, and refined
the most promising incumbents. Shortlisted candidates were re-evaluated in the
root process before acceptance, and the test interval was not accessed during
this search. Models were ranked by held-out $E_{\mathrm{data}}$. Improvement relative to a
model's own classical baseline was defined as
$100(E_{\mathrm{data}}^{\mathrm{base}}-E_{\mathrm{data}}^{\mathrm{model}})
/E_{\mathrm{data}}^{\mathrm{base}}$.

\paragraph{Results}
Validation selected the
following baseline-specific structures:
\begin{align}
    g_{\mathrm{G}}(\rho)
        &= \exp\!\left(c_0+a\rho+b\rho^2+dC(\rho)\right), \\
    g_{\mathrm{IDM}}(\rho)
        &= \exp\!\left(aC(\rho)+bH_2(\rho)C(\rho)\right), \\
    g_{\mathrm{W}}(\rho)
        &= \exp\!\left(a\rho\left(1-\rho/\rho_{\max}\right)\right), \\
    g_{\mathrm{T}}(\rho)
        &= \exp\!\left(a\rho C(\rho)\right), \\
    g_{\mathrm{DC}}(\rho)
        &= \exp\!\left(c_0+aS_3(\rho)+bC(\rho)\right),
    \label{eq:traffic-selected-corrections}
\end{align}
with
\begin{equation}
    S_3(\rho)=\operatorname{conv}_3(\rho,\mathbf{1}),
    \qquad
    C(\rho)=S_3(\rho)-3\operatorname{conv}_1(\rho,\mathbf{1}),
\end{equation}
where $C$ is a local density-contrast feature, and let $H_2(\rho)$ denote the
quadratic Hodge-star feature obtained by squaring the dual one-cochain
$\star\rho$ and mapping it back to primal nodes. 
These structures admit a direct physical reading. Because every $g$ is a
positive multiplier, a correction changes the magnitude of the baseline flux
and velocity without changing the baseline's zero-flow states. The contrast
$C$ vanishes in a spatially homogeneous state, so terms containing $C$ act
only where the density varies over the short convolution window. They can
therefore be interpreted as compact non-equilibrium corrections at developing
traffic fronts, rather than as changes to the homogeneous fundamental diagram.

The refitted Greenshields coefficients have $c_0>0$ and $a,b,d<0$. The
constant term raises the low-density branch, the negative linear and quadratic
terms progressively temper this uplift as density increases, and the contrast
term adjusts the flux near spatial transitions. This supplements the rigid
parabolic Greenshields curve with both density-dependent curvature and a local
traffic-state correction. For IDM, both fitted coefficients are negative and
the correction is proportional to $C$; the non-negative Hodge feature $H_2$
modulates the strength of that response with local traffic loading. The model
therefore leaves uniform flow unchanged but reacts more strongly to density
variations in heavily loaded states, consistent with a queue-front or
anticipation effect absent from an equilibrium closure.

The Weidmann correction is the most directly interpretable. Its fitted
$a>0$, so it increases flux at intermediate densities while remaining exactly
one at both $
ho=0$ and $
ho=\rho_{\max}$. It preserves the free-flow and jam
limits and modifies only the interior of the diagram. The Triangular model has
$a<0$ and gates the contrast by density: it is inactive in empty or homogeneous
traffic but becomes more influential as congestion develops. Finally, the Del
Castillo fit has $c_0>0$ and $a,b<0$. It combines a small global uplift with
suppression as the three-cell density aggregate $S_3$ grows and an additional
response at spatial density contrasts. Taken together, the selected structures
suggest that the residual bias is baseline dependent: some diagrams mainly
need a correction to their homogeneous density dependence, whereas others
benefit from sensitivity to short-range traffic heterogeneity. These
interpretations are structural hypotheses, not identified driver-behaviour
mechanisms, and require validation on other roads and traffic regimes.

\begin{figure}[t]
    \centering
    \includegraphics[width=\textwidth]
        {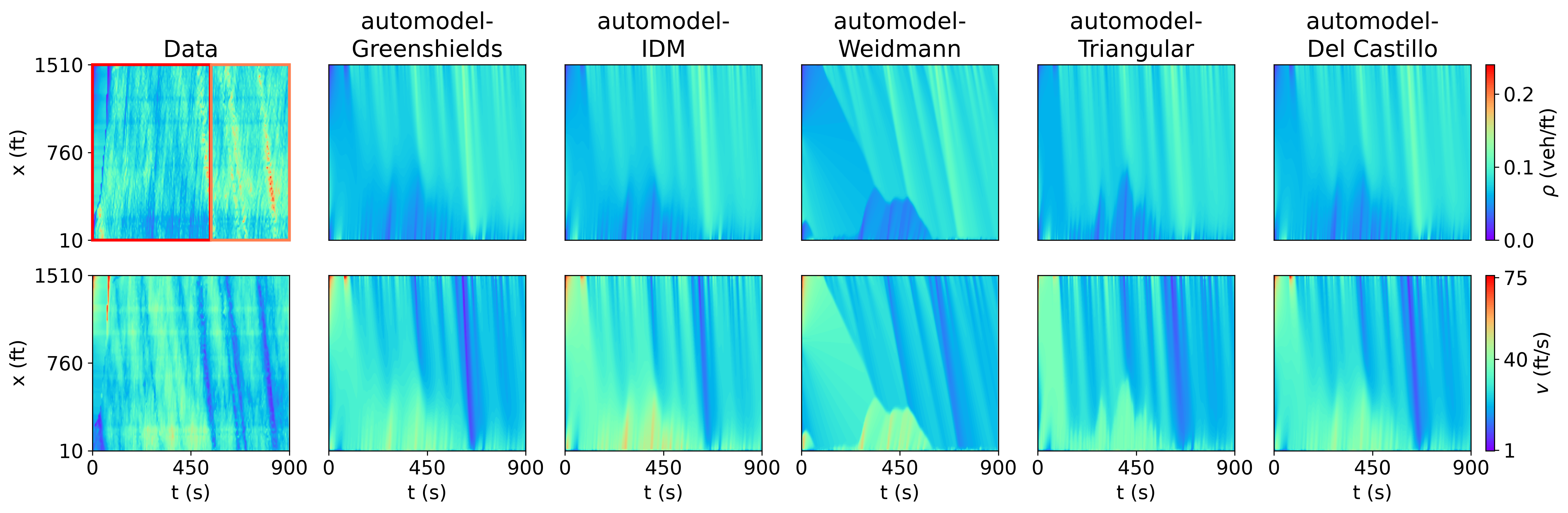}
    \caption{Measured I80 fields and predictions from the five selected
    \automodel{} corrections. Density is shown in the top row and velocity in the
    bottom row; the first column contains the measurements and the remaining
    columns the model predictions. The outlined regions in the measured
    density panel distinguish the development and held-out time intervals.}
    \label{fig:traffic-fields}
\end{figure}

Table~\ref{tab:traffic-test-results} compares the five classical diagrams with
the corresponding structures found in the \automodel{} search.
\automodel{}--Del
Castillo attained the lowest held-out density, velocity, and aggregate errors,
with $E_{\mathrm{data}}=6.780$. The \automodel{}--Greenshields correction produced
the largest improvement over its baseline, reducing test $E_{\mathrm{data}}$
by 23.17\%. \automodel{} improved held-out $E_{\mathrm{data}}$ for four of the five
baseline families; the exception was IDM, for which the score increased by
2.30\%. Figure~\ref{fig:traffic-fields} shows the measured spatio-temporal
fields beside the five \automodel{} predictions. The models reproduce the broad
traffic regimes but smooth out part of the fine-scale variability in the data.

\begin{table}[t]
    \centering
    \small
    \setlength{\tabcolsep}{3.5pt}
    \caption{Held-out I80 prediction errors. Lower is better for all error
    columns. Parentheses give rank by full-precision test
    $E_{\mathrm{data}}$ across the ten models. $\Delta_{\mathrm{base}}$ is the
    percentage reduction in test $E_{\mathrm{data}}$ relative to the model's
    own classical baseline; positive values indicate improvement. AM denotes
    \automodel{}.}
    \label{tab:traffic-test-results}
    \begin{tabular}{@{}lrrrr@{}}
        \toprule
        Model & $E_\rho$ & $E_v$ & $E_{\mathrm{data}}$ (rank)
              & $\Delta_{\mathrm{base}}$ \\
        \midrule
        Greenshields       & 0.294 & 0.313 & 9.230 (10) & --         \\
        AM--Greenshields   & 0.255 & 0.277 & 7.092 (3)  & $+23.17\%$ \\
        IDM                & 0.258 & 0.270 & 6.957 (2)  & --         \\
        AM--IDM            & 0.256 & 0.277 & 7.117 (4)  & $-2.30\%$  \\
        Weidmann           & 0.265 & 0.288 & 7.651 (8)  & --         \\
        AM--Weidmann       & 0.265 & 0.281 & 7.459 (7)  & $+2.51\%$  \\
        Triangular         & 0.269 & 0.294 & 7.919 (9)  & --         \\
        AM--Triangular     & 0.256 & 0.277 & 7.118 (5)  & $+10.11\%$ \\
        Del Castillo       & 0.257 & 0.279 & 7.197 (6)  & --         \\
        AM--Del Castillo   & \textbf{0.253} & \textbf{0.267}
                           & \textbf{6.780 (1)} & $+5.80\%$ \\
        \bottomrule
    \end{tabular}
\end{table}

\section{Discussion}

Across the five case studies, general-purpose coding agents equipped with \automodel{} were able to discover useful and scientifically plausible candidate models, ranging from standalone analytical expressions to equations embedded in simulators.
Although the skill also allows neural architecture changes in principle, these case studies do not evaluate neural architecture search.
In astrophysics, the selected stellar mass function improved on the tested literature forms while extending discovery beyond the complexity ceiling accessible to Exhaustive Symbolic Regression.
In precision agriculture, the discovered expressions outperformed traditional baselines on nitrogen response curve fitting and performed competitively with a specialized neural SR approach.
In automotive modeling, an asymmetric hyperbolic-tangent formulation provided an alternative to the unshifted Pacejka's Magic Formula, retaining an accuracy advantage at a previously unused load.

The cardiovascular and traffic applications further illustrate how discovery can proceed through targeted modifications of established models.
In CircAdapt, a shared modification of myocardial activation was evaluated while the existing patient-specific parameters were independently recalibrated.
In traffic modeling, compact multiplicative corrections retained the sign and zero-flow states of classical fundamental diagrams while introducing additional density dependence or sensitivity to spatial heterogeneity.
These examples show that agent-guided discovery can build on existing scientific knowledge and software, allowing researchers to preserve selected properties while investigating where the original formulation is inadequate. This customisability and interactivity with the user are lacking in most function discovery algorithms.

The common element across these applications is the discovery procedure, whose implementation was adapted to each problem.
The researcher still defines the scientific objective, admissible changes, fitting procedure, and evidence required to accept a result.
The agent can then inspect the implementation, propose structural hypotheses, conduct experiments, and use their outcomes to guide subsequent attempts under the user's supervision.
Interactivity is particularly useful when inspecting failures reveals an unsuitable functional family or exposes a trade-off between accuracy, complexity, and interpretability, as occurred in precision agriculture.
In the cardiovascular application, a proposed modification passed the identity check, but a researcher-led audit showed that its new coefficient duplicated the effect of an existing parameter.
The resulting workflow therefore supports both model search and the practical work of formulating and checking a modeling problem.

Model discovery algorithms are often compared in terms of the compute or runtime required to reach a given level of accuracy~\cite{la2021contemporary}.
Comparisons involving \automodel{} should account for numerical calibration, agent inference, and researcher involvement.
Stochasticity can be assessed through repeated independent runs, while the amount and timing of human feedback should also be documented.
The most direct comparison is with the same coding agent working without the skill. As a first test of this, in the astrophysics case study of Sec.~\ref{sec:astro} we gave three independent sessions of the same agent and model the evaluator, data, and already-evaluated functions used by \automodel{} but replacing the skill with a generic instruction to improve the model. Their wall time and compute were capped at those of the \automodel{} run and their token budget at our estimate of its usage. Each session considered the task complete after 40--65 minutes, having used less than half of its token budget. Although two of the three found functions that beat A5 on one of the two datasets, only one found a new physically admissible function beating the Bernardi function on both datasets and by significantly smaller margins than A5. The best \automodel{}-found function thus remained clearly the best overall; this appears to be a consequence of the skill's enforced meta-iterations continuing the search until a single function succeeded on both datasets, thus increasing the agents' persistence. More systematic control tests---and more detailed comparisons with traditional search algorithms---remain to be made.

Finally, accurate and physically admissible expressions do not by themselves establish an underlying mechanism.
The discovered structures provide hypotheses whose mechanistic interpretation requires independent validation.
Controlled experiments with known governing equations could additionally assess recovery of the generating structure, accounting for mathematically equivalent formulations and limits of identifiability. In real-world applications, this requires the insight of domain experts.

\section{Conclusion}

We introduced \automodel{}, a portable skill for scientific model discovery with general-purpose coding agents.
Across five case studies in different scientific fields, we showcased the effectiveness of a shared workflow that combines domain expertise, numerical experimentation, and agent-guided exploration to discover useful models and refine existing scientific formulations.
\automodel{} provides a reusable foundation for extending this approach to new modeling problems and scientific domains.
The skill lives as an evolving, open source project at \url{https://github.com/unlayer-ai/automodel}: contributions are welcome.

\section*{Acknowledgments}
HD is supported by Royal Society University Research Fellowship 211046.

EB is supported by a scholarship funded by Ministerial Decree no.~118 of 2 March 2023, under the National Recovery and Resilience Plan (NRRP), financed by the European Union --- NextGenerationEU --- Mission 4, Investment 4.1.

GM and JWS were supported by a USDA-NIFA-AFRI project, titled “Using Precision Technology in On-farm Field Trials to Enable Data-Intensive Fertilizer Management” (Accession Number 2016-68004-24769), and by a USDA-NRCS program, titled “Improving the Economic and Ecological Sustainability of US Crop Production through On-Farm Precision Experimentation” (Award Number NR213A7500013G021).

The work of AL is supported by the European Union (European Research Council (ERC), ALPS, 101039481). Views and opinions expressed are however those of the author(s) only and do not necessarily reflect those of the European Union or the ERC Executive Agency. Neither the European Union nor the granting authority can be held responsible for them.

\section*{Author Contributions}

MV conceived the project, formulated the problem and overall methodology, and led the manuscript preparation.
MV and HD developed the \automodel{} skill.  
HD conducted and wrote the astrophysics case study (Sec.~\ref{sec:astro}). 
EB conducted and wrote the cardiovascular modeling case study (Sec.~\ref{sec:use-case-2}). 
GM and JWS conducted and wrote the precision agriculture case study (Sec.~\ref{sec:use-case-3}). 
LR conducted and wrote the automotive tire modeling case study (Sec.~\ref{sec:use-case-4}). 
AL conducted and wrote the traffic flow modeling case study (Sec.~\ref{sec:use-case-5}). 
All authors contributed to reviewing, editing and approving the final manuscript.

\bibliographystyle{unsrt}  
\bibliography{references}

\end{document}